\documentclass[letterpaper]{article} 
\usepackage{aaai21}  
\usepackage{times}  
\usepackage{helvet} 
\usepackage{courier}  
\usepackage[hyphens]{url}  
\usepackage{graphicx} 
\usepackage{natbib}  
\usepackage{caption} 
\usepackage{tabularx}

\newcommand{\dc}{Direct Connections}
\newcommand{\cdc}{Contextual  Direct  Connections}
\newcommand{\hop}{Neighborhood}

\usepackage{sidecap}

\usepackage{latexsym}
\usepackage{soul}
\usepackage{amsfonts}
\usepackage{amsmath}
\usepackage{amssymb}
\usepackage{mathptmx}
\usepackage{nicefrac}
\usepackage{booktabs} 
\usepackage{makecell} 
\usepackage{graphicx}
\usepackage[inline]{enumitem}
\usepackage[font=small]{subcaption}
\usepackage[font=small]{caption}
\usepackage{xargs} 

\usepackage[autostyle]{csquotes}  

\usepackage[T1]{fontenc}

\usepackage{color}
\usepackage{soul}
\usepackage[textsize=scriptsize]{todonotes}
\definecolor{pigment}{rgb}{0.2, 0.2, 0.6}

\definecolor{blue}{RGB}{0, 93, 170}			
\definecolor{darkgreen}{HTML}{3bb35b}

\newcommand{\twc}{\texttt{TextWorld Commonsense}}
\newcommand{\twcshort}{\texttt{TWC}}

\definecolor{darkyellow}{HTML}{f0b71d}

\newcommand{\ignore}[1]{}
 \usepackage{multirow}
 \usepackage{float}

\title{Text-based RL Agents with Commonsense Knowledge:\\New Challenges, Environments and Baselines}
\author{
        Keerthiram Murugesan\textsuperscript{\rm 1},
        Mattia Atzeni\textsuperscript{\rm 2},
        Pavan Kapanipathi\textsuperscript{\rm 1},
        Pushkar Shukla\textsuperscript{\rm 3},
        Sadhana Kumaravel\textsuperscript{\rm 1},
        Gerald Tesauro\textsuperscript{\rm 1},
        Kartik Talamadupula\textsuperscript{\rm 1},
        Mrinmaya Sachan\textsuperscript{\rm 4},
        Murray Campbell \textsuperscript{\rm 1}
}
\affiliations{
    \textsuperscript{\rm 1}IBM Research, Yorktown Heights~
    \textsuperscript{\rm 2}IBM Research, Zurich ~ 
    \textsuperscript{\rm 3}TTI Chicago ~ 
    \textsuperscript{\rm 4}ETH Zurich\\
    \texttt{keerthiram.murugesan@ibm.com},~ \texttt{atz@zurich.ibm.com},~ \texttt{kapanipa@us.ibm.com},~ \\
    \texttt{pushkarshukla@ttic.edu},~ \texttt{sadhana.kumaravel1@ibm.com},~ \texttt{gtesauro@us.ibm.com},~\\
    \texttt{krtalamad@us.ibm.com},~ \texttt{mrinmaya.sachan@inf.ethz.ch},~ \texttt{mcam@us.ibm.com}
}

\begin{document}

\maketitle

\begin{abstract}

Text-based games have emerged as an important test-bed for Reinforcement Learning (RL) research, requiring RL agents to combine grounded language understanding with sequential decision making.
In this paper, we 
examine the problem of infusing RL agents with commonsense knowledge. 
Such knowledge would allow agents to efficiently act in the world by pruning out implausible actions,
and 
to perform look-ahead planning to determine how current actions might affect future world states.
We design a new text-based gaming environment called \twc\ (\twcshort) for training and evaluating RL agents with a specific kind of commonsense knowledge about objects, their attributes, and affordances. We also introduce several baseline RL agents which track the sequential context and dynamically retrieve the relevant commonsense knowledge from ConceptNet. 
We show that agents which incorporate commonsense knowledge in \twcshort\ perform better, while acting more efficiently. We conduct user-studies to estimate human performance on \twcshort\ and show that there is ample room for future improvement.

\end{abstract}


\section{Introduction}
\label{sec:intro}

Over the years, simulation environments have been used extensively to drive advances in reinforcement learning (RL). A recent framework that has received much attention is TextWorld ({\tt TW})~\cite{cote18textworld}, where an agent must interact with an external environment to achieve a given goal using only the modality of text. TextWorld and similar text-based environments seek to bring advances in grounded language understanding to a sequential decision making setup.


\begin{figure}[tp]
    \centering
    \includegraphics[width=\linewidth]{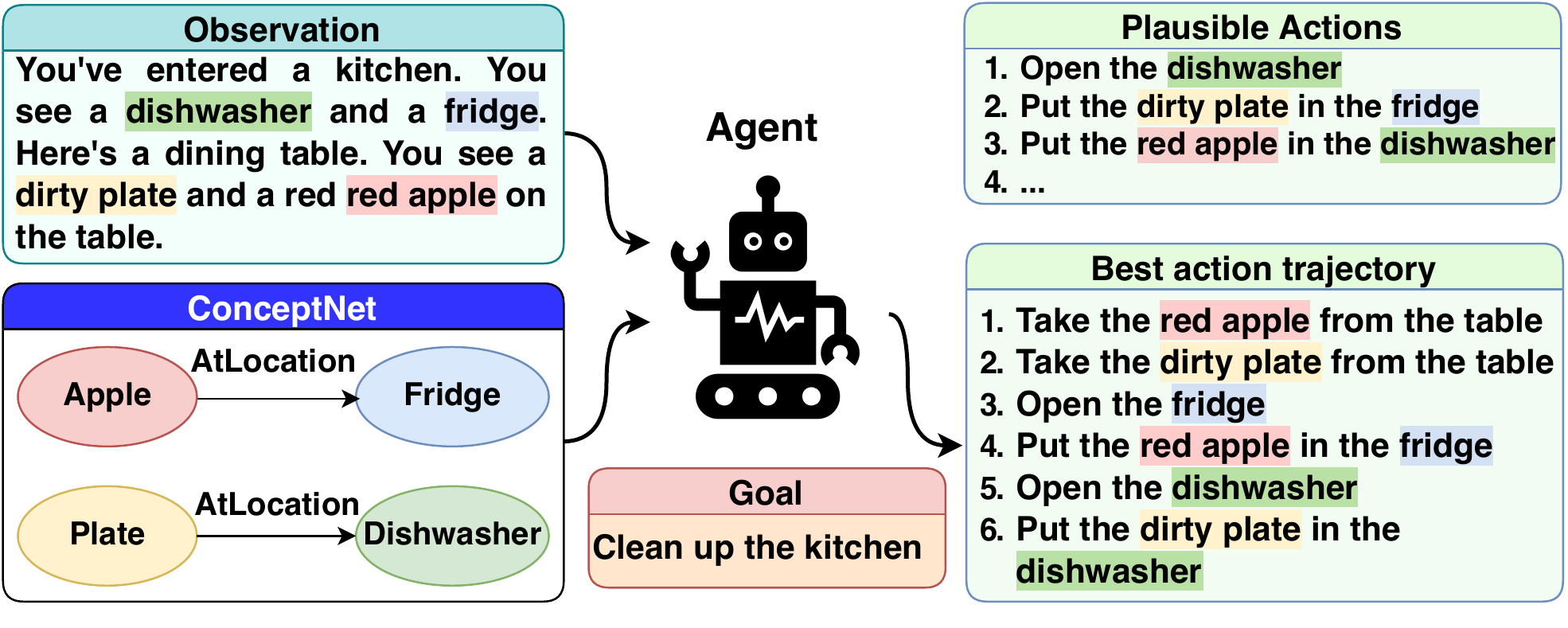}
    \caption{{
    Illustration of a \twcshort\ game. 
    The agent is given an initial observation (top left) and has to produce the list of actions (bottom right) that are necessary to achieve the goal (bottom center) using relevant commonsense knowledge from ConceptNet (bottom left).
    }}
    \label{fig:kitchen_cleanup}
    \vspace{-5mm}
\end{figure}

While existing text-based games are valuable for RL research, they fail to test a key aspect of human intelligence: common sense. Humans capitalize on commonsense (background) knowledge about entities -- properties, spatial relations, events, causes and effects, and other social conventions  -- while interacting with the world \cite{Mccarthy1960ProgramsWC,winograd1972understanding,davis2015commonsense}.
Motivated by this, we propose a novel text-based environment called \twc\ (or \twcshort), where the agent is expected to use commonsense knowledge stored in knowledge bases such as {\em ConceptNet} \cite{liu2004conceptnet,speer2017conceptnet} to act efficiently.
\twcshort\ is a sandbox environment similar to {TextWorld} where the agent has to clean up a house. Achieving goals in this environment requires commonsense knowledge about objects, their properties, locations, and affordances.
Efficient use of commonsense knowledge would allow the agent to select correct and applicable actions at each step: i.e., improve sample efficiency by reducing exploration. Moreover, commonsense knowledge would help the agent to perform look-ahead planning and determine how current actions might affect future world states \cite{juba2016integrated}.
Fig~\ref{fig:kitchen_cleanup} presents a running example from \twcshort\ that illustrates how the agent can leverage a commonsense knowledge base (KB). 


Validating such environments is challenging, and requires: (1) verifying the information used in the games; (2) evaluating baseline agents that are capable of utilizing external commonsense knowledge against counterparts that do not; and (3) providing empirical evidence to show that the environment can drive future research. In this work, we address each of these by first performing human annotations to validate the correctness and completeness of the \twcshort\ environment. Next, we design a framework of agents that combine text-based agents with commonsense knowledge. The agents can dynamically retrieve relevant knowledge from a commonsense KB. Finally, based on human performance on the generated games and manual selection of commonsense knowledge, we discuss and justify the importance of such an environment in driving future research.

\noindent {\bf Contributions}: The main contributions of this paper are the following: (1) we propose an novel environment called \twcshort\ to evaluate the use of commonsense knowledge by RL agents; (2) we introduce baselines that use commonsense knowledge from ConceptNet and show that common sense indeed helps in decision making;
(3) whereas our model with common sense performs well,
we show a pronounced gap in performance between automated agents and humans in the \twcshort\ environment. This substantiates our claim that \twcshort{} provides a challenging test-bed for RL agents and can act as a spur to further research in this area.


\section{TextWorld Commonsense (\twcshort)}
\label{sec:textworld_commonsense}


Existing text-based games \cite{adhikari2020learning,cote18textworld} severely restrict the amount and variety of commonsense knowledge that an agent needs to know and exploit.
Thus, in this paper, we create and present a new domain -- \twc\ (\twcshort) -- by reusing the TextWorld \cite{cote18textworld} engine in order to generate text-based environments where RL agents need to effectively retrieve and use commonsense knowledge.
Commonsense can be defined very broadly and in various ways \cite{fulda2017whatcanyoudo}. In this paper, we mainly focus on commonsense knowledge that pertains to objects, their attributes, and affordances\footnote{Gibson in his seminal work~\cite{gibson1978ecological} refers to affordance as ``properties of an object [...] that determine what actions a human can perform on them''. 
}.

\subsection{Constructing \twcshort}
\label{subsec:constructing_twc}

We built the \twcshort\ domain as a house clean-up environment where the agent is required to obtain knowledge about typical objects in the house, their properties, and expected location from a commonsense knowledge base.
The environment is initialized with random placement of objects in various locations.
The agent's high level goal is to tidy up the house by putting objects in their {\tt commonsense} locations. This high level goal may consist of multiple sub-goals requiring commonsense knowledge. For example, for the sub-goal: \textit{put the apple inside the refrigerator}, commonsense knowledge from ConceptNet such as \small (\texttt{Apple}  $\rightarrow$ \texttt{AtLocation}  $\rightarrow$ \texttt{Refrigerator})
\normalsize can assist the agent.

\vspace{1mm}

\noindent {\bf Goal Sources}: While our main objective was to create environments that require commonsense, we did not want to bias \twcshort{} towards any of the existing knowledge bases. We additionally wanted to rule out the possibility of data leaks in situations where both the environment as well as the external knowledge came from the same part of a specific commonsense knowledge base (KB) like ConceptNet. For the construction of the \twcshort\ goal instances, we picked sources of information that were orthogonal to existing commonsense KBs. Specifically, we used: (1) the {\em picture dictionary from 7ESL}\footnote{\url{https://7esl.com/picture-dictionary}}; (2) the {\em British Council's vocabulary learning page}\footnote{\url{https://learnenglish.britishcouncil.org/vocabulary/beginner-to-pre-intermediate}
}; (3) the {\em English At Home vocabulary learning}
page\footnote{\url{https://www.english-at-home.com/vocabulary}}; and (4) {\em ESOL courses}\footnote{\url{https://www.esolcourses.com/topics/household-home.html}}.
We collected vocabulary terms from these sources and manually aggregated this content in order to build a
dataset that lists several kinds of objects that are typically found in a house environment. For each object, the dataset specifies a list of plausible and coherent locations.

\vspace{1mm}

\noindent {\bf Instance Construction}: A \twcshort\ instance is sampled from this dataset, which includes a configuration of $8$ room types and a total of more than $900$ entities (Table~\ref{tab:twc_numbers_examples}). The environment includes three main kinds of entities: objects, supporters, and containers. Objects are entities that can be carried by the agent, whereas supporters and containers are furniture where those objects can be placed. 
Let {\tt o} represent the object or entity in the house; {\tt r} represent the room that the entity is typically found in; and {\tt l} represent the location inside that room where the entity is typically placed.
In our running example, {\tt o:apple} is an entity, {\tt l:refrigerator} is the container, and {\tt r:kitchen} is the room.
Via a  manual verification process (which we elucidate next in Section~\ref{subsec:verifying_twc_annotations}) we ensure that the associations between entities, supporters/containers, and rooms reflect commonsense.
As shown in Table~\ref{tab:twc_numbers_examples}, we collected a total of $190$ objects from the aforementioned resources. We further expanded this list by manually annotating the objects with qualifying properties, which are usually adjectives from a predefined set (e.g., a shirt may have a color and a specific texture). This allows increasing the cardinality of the total pool of objects for generating \twcshort\ environments to more than $800$.

\begin{table}[]
\centering
\resizebox{1.0\linewidth}{!}{%
\centering
\begin{tabular}{|c|c|c|}
\hline
 & \textbf{Count} & \textbf{Examples} \\ \hline
\textbf{Rooms} & 8 & \begin{tabular}[c]{@{}c@{}}\textit{kitchen}, \textit{backyard}\end{tabular} \\
\hline
\textbf{Supporters/Containers} & 56 & \begin{tabular}[c]{@{}c@{}}\textit{dining table}, \textit{wardrobe}\end{tabular} \\
\hline
\textbf{Unique Objects} & 190 & \begin{tabular}[c]{@{}c@{}}\textit{plate}, \textit{dress}\end{tabular} \\
\hline
\textbf{Total Objects} & 872 & \begin{tabular}[c]{@{}c@{}}\textit{dirty plate}, \textit{clean red dress}\end{tabular} \\
\hline
\textbf{Total Entities} & 928 & \begin{tabular}[c]{@{}c@{}}\textit{dirty plate}, \textit{dining table}\end{tabular} \\
\hline
\end{tabular}%
}
\caption{{Statistics on the number of entities, supporters/containers, and rooms in the \twcshort\ domain.}}
\label{tab:twc_numbers_examples}
\vspace{-3mm}
\end{table}




\begin{figure*}[ht]
    \centering
    \includegraphics[width=\textwidth]{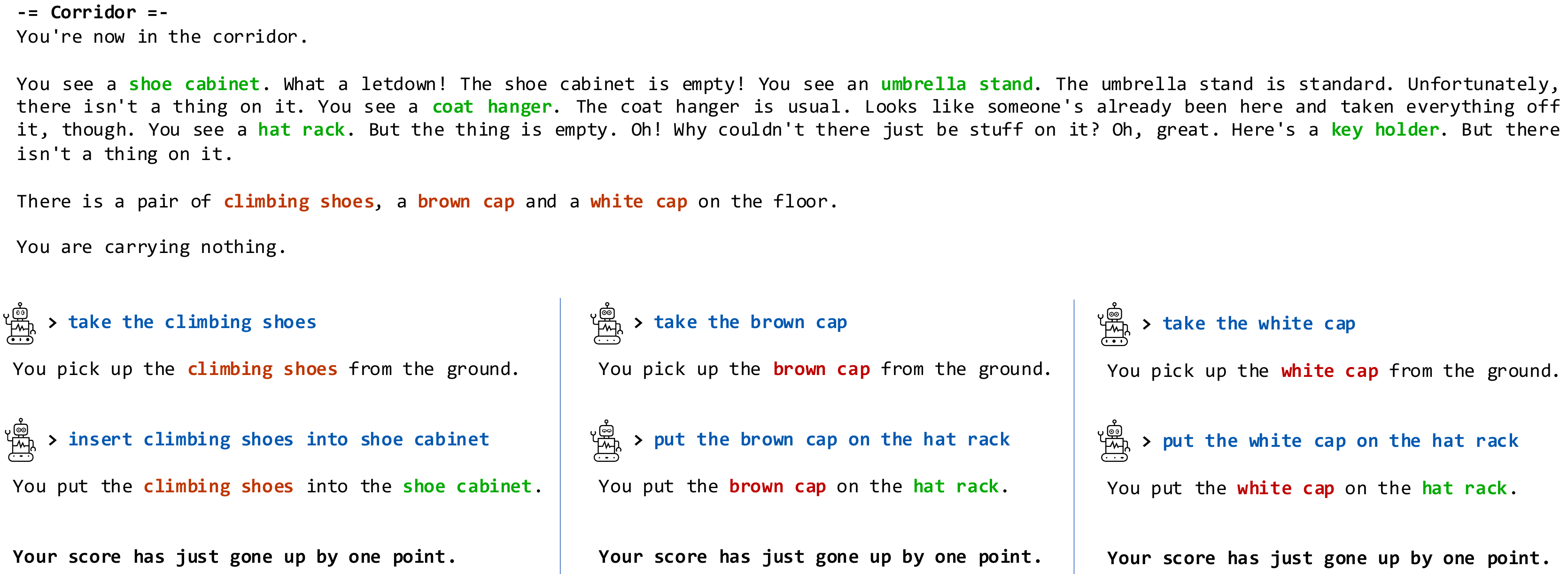}
    \caption{Sample game walkthrough for a game with \emph{medium} difficulty level. Best viewed in colors. Highlights are not available to the agents and are shown for illustrative purpose only.}
    \label{fig:medium_game_walkthrough}
\end{figure*}

\subsection{Verifying \twcshort}
\label{subsec:verifying_twc_annotations}

\begin{table}[]
\centering
\resizebox{1.0\linewidth}{!}{%
\small
\begin{tabular}{|c|c|c|}
\hline
 & \textbf{Correctness} & \textbf{Completeness} \\\hline
\textbf{Rated Commonsense} & 669 & 47 \\\hline
\textbf{Rated NOT Commonsense}  & 31 & 253 \\\hline
\end{tabular}%
}
\caption{{Statistics from the human annotations to verify \twcshort}}
\label{tab:agreement}
\vspace{-5mm}
\end{table}

In order to ensure that \twcshort\ reflects commonsense knowledge, we set up two annotation tasks to verify the environment goals (i.e., goal triples of the form $\langle$\texttt{o}, \texttt{r}, \texttt{l}$\rangle$, where \texttt{o} denotes the object, \texttt{r} denotes a room, and \texttt{l} a location within that room, as defined in Section \ref{subsec:constructing_twc}). 
The first task is meant to verify the correctness of the goals and evaluate whether the goal $\langle$\texttt{o}, \texttt{r}, \texttt{l}$\rangle$ triples make sense to humans. The second task is aimed at verifying completeness, i.e. that other triples in the environment do not make sense to humans.

\vspace{1mm}

\noindent{\bf Verifying Correctness:} To test the correctness of our environments, we
asked our human annotators to determine whether they would consider a given room-location combination in the goal $\langle$\texttt{o}, \texttt{r}, \texttt{l}$\rangle$ to be a reasonable place for the object \texttt{o}. If so, the instance was labeled as positive, and as negative otherwise. We collected annotations from $10$ annotators, across a total of $205$ unique $\langle$\texttt{o}, \texttt{r}, \texttt{l}$\rangle$ triples. Each annotator labeled $70$ of these triples, and each triple was assigned to at least $3$ distinct annotators. The annotators were not given any other biasing information, and all annotators worked independently.
We show the overall agreement of the annotators with \twcshort's goals in Table \ref{tab:agreement}. The high agreement from the annotators demonstrates that the goal $\langle$\texttt{o}, \texttt{r}, \texttt{l}$\rangle$ triples reflect human commonsense knowledge.

\begin{table}[h]
\center
\begin{tabular}{|l|l|l|l|}
\hline
 & \textbf{\#objects} & \textbf{\#Objects to find} & \textbf{\#Rooms} \\ \hline
\textbf{Easy} & 1 & 1 & 1 \\ \hline
\textbf{Medium} & 2, 3 & 1, 2, 3 & 1 \\ \hline
\textbf{Hard} & 6, 7 & 5, 6, 7 & 1, 2 \\ \hline
\end{tabular}%
\caption{{Specification of \twcshort{} games}}
\label{tab:difficulty_levels}
\vspace{-5mm}
\end{table}

\vspace{1mm}

\noindent{\bf Verifying Completeness:} Similar to the above annotation exercise, we also asked human annotators to determine if a non-goal $\langle$\texttt{o}, \texttt{r}, \texttt{l}$\rangle$ triple made sense to them. In addition to the $70$ triples mentioned above, each of the $M = 10$ annotators were asked to label as either positive or negative a set of $30$ non-goal triples. In order to provide annotators with an informative set of non-goal $\langle$\texttt{o}, \texttt{r}, \texttt{l}$\rangle$ triples, we used GloVe~\cite{pennington2014glove} to compute embeddings for each location in \twcshort{}. For a given object \texttt{o}, a non-goal location \texttt{l}' was then selected among those most similar to the goal location \texttt{l}, according to the cosine similarity between the embeddings of \texttt{l} and \texttt{l}'. As before, each non-goal triple was assigned to at least $3$ annotators from a set that comprises a total of $97$ triples. As we see in Table \ref{tab:agreement}, the annotators seldom find a hypothesized non-goal $\langle$\texttt{o}, \texttt{r}, \texttt{l}$\rangle$ triple as commonsensical.

\vspace{1mm}

\noindent{\bf Annotator Reliability:}
For our overall annotation exercise, we can report inter-annotator agreement statistics, as the overall annotation is no longer imbalanced in terms of label marginals. We report a {\em Krippendorff's alpha}~\cite{krippendorff2018content} $\alpha_{\kappa} = 0.74$. This number is over the accepted range for agreement and shows that our annotators have strong agreement when rating the triples.

\subsection{Generating \twcshort{} Games}
\label{sec:twc_games}

We used the TextWorld engine to build a set of text-based games where the goal is to tidy up a house by putting objects in the goal locations specified in the aforementioned \twcshort{} dataset. The games are grouped into three difficulty levels (easy, medium, and hard) depending on the total number of objects in the game, the number of objects that the agent needs to find (the remaining ones are already carried by the agent at the beginning of the game) and the number of rooms to explore. The values of these properties are randomly sampled from the ones listed in Table \ref{tab:difficulty_levels}.
For each difficulty level, we provide a training set and two test sets. The training sets were built out of $\frac{2}{3}$ of the unique objects reported in Table \ref{tab:twc_numbers_examples}. For the first test set, we used the same set of objects as the training games. We call this set the \textit{in} distribution test set. For the second test set, we employed the remaining $\frac{1}{3}$ objects to create the evaluation games. We call this set the \textit{out} of distribution test set. This allows us to investigate not only the capability of the agents to generalize within the same distribution of the training data, but also their ability to achieve generalization to unseen entities. Figure \ref{fig:medium_game_walkthrough} shows a game walkthrough for a specific game in the medium difficulty level.

\subsection{Benchmarking Human Performance}
\label{subsec:human_performance_twc}
To complete our benchmarking of the \twcshort\ domain, we conducted yet another human annotation task, focusing on the performance of human game-players. Such an experiment is essential to establishing the performance of human players, who are generally regarded as proficient at exploiting commonsense knowledge. We set up an interactive interface to \twcshort\ via a Jupyter notebook, which was then used by players to interact with the same games that we evaluated all the other RL agents on. We recorded all moves (steps) made by players, as well as the reward collected. At each step, the players were shown the current context of the game in text format, and given a drop-down box with the full list of possible actions. Once the player picked an action, it was executed; and this process repeated until all possible goals in the game had been accomplished. A total of $16$ annotators played $104$ instances of \twcshort\ games, spread across the {\em easy}, {\em medium}, and {\em hard} levels. Each difficulty level had $5$ games, each from the train and test distributions, for a total of $30$ unique games. Each unique game was annotated by a minimum of $3$ annotators. 
The results are presented in Table~\ref{tab:generalization_res}, along with the experimental results in Section~\ref{sec:experiments}, to allow for direct comparison with the \twcshort\ agents.

\vspace{1mm}

\section{\twcshort\ Agents}
\label{sec:textworld_as_pomdp}

\begin{figure}
    \centering
    \includegraphics[width=1.0\linewidth]{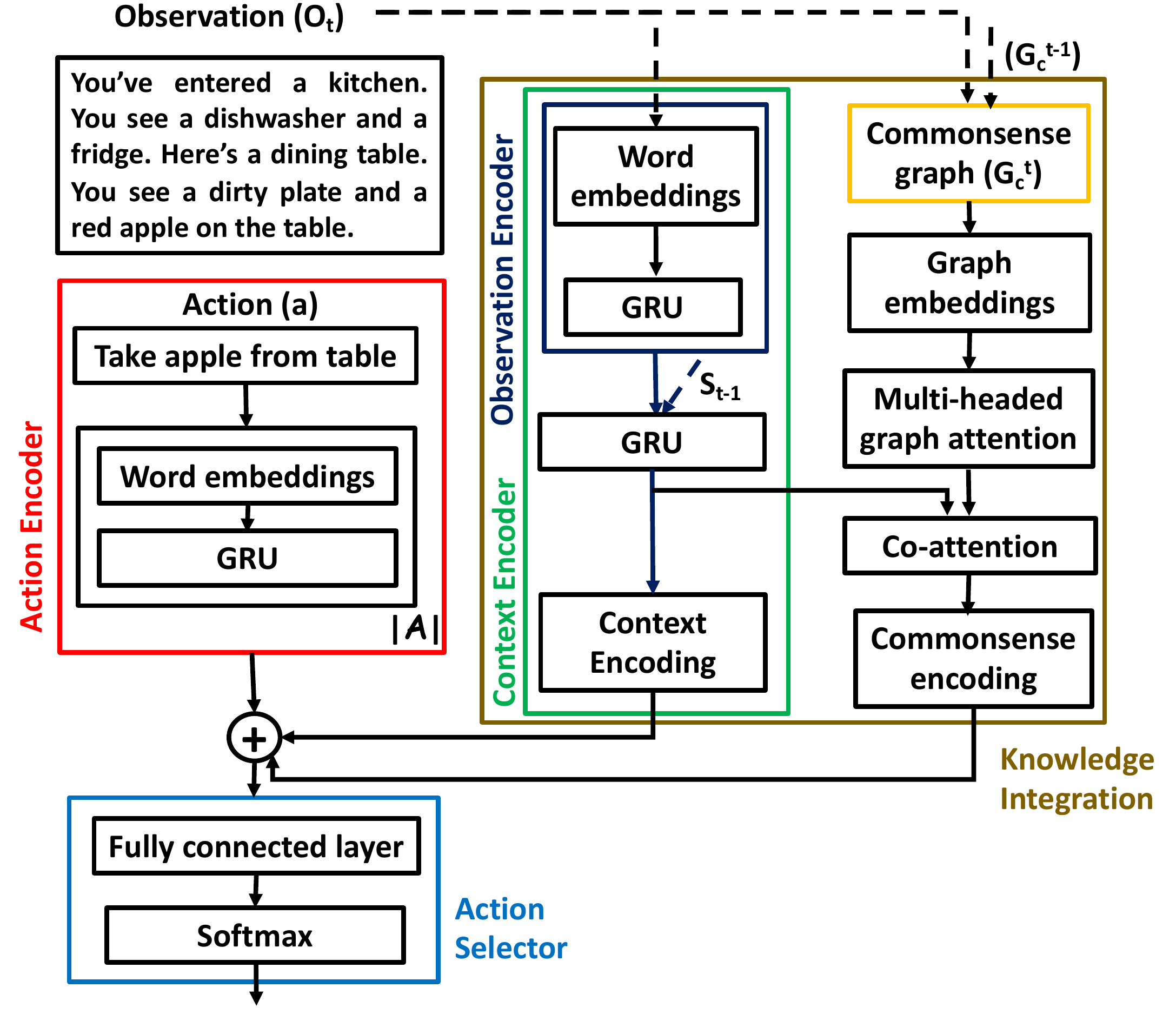}
    \caption{Overview of our framework's decision making at any given time step. The framework comprises of the following components (visually shown in color): (a) \textcolor{red}{action encoder} which encodes all admissible actions $a \in \mathcal{A}$, (b) \textcolor{pigment}{observation encoder} which encodes the observation $o_t$, (c) \textcolor{darkgreen}{context encoder}, which encodes the dynamic context $C_t$, (d) a dynamic \textcolor{darkyellow}{common sense subgraph} of ConceptNet $G_C^t$ extracted by the agent, (e) a \textcolor{brown}{knowledge integration} component, which combines the information from textual observations and the extracted common sense subgraph, and (f) an \textcolor{cyan}{action selection} module. $\oplus$ denotes the concatenation operator.}
    \label{fig:block_diagram}
    \vspace{-5mm}
\end{figure}

Text-based games can be seen as partially observable Markov decision processes (POMDP)~\cite{kaelbling1998planning} where the system dynamics are determined by an MDP, but the agent cannot directly observe the underlying state.
The agent receives a reward at every time step
and its goal is to maximize the expected discounted sum of rewards.
The \twcshort{} games allow the agent to perceive and interact with the environment via text. Thus, the observation at time step $t$, $o_t$, is presented as a sequence of tokens ($o_t =  \{o_t^1, \dots o_t^N\}$). Similarly, each action $a$ is also denoted as a sequence of tokens $\{a^1, \dots, a^M\}$.
The goal of this project is to test RL agents with commonsense. Hence, the agents also have access to a commonsense knowledge base; and are allowed to use it while selecting actions.
To model \twcshort, we design a framework that can: (a) learn representations of various actions; (b) learn from sequential context; (c) dynamically retrieve the relevant commonsense knowledge; (d) integrate the retrieved commonsense knowledge with the context; and (e) predict next action. A block diagram of the framework is shown in Fig \ref{fig:block_diagram}. We describe the various components of our framework below.


\subsection{Action and Observation Encoder}
\label{sec:input_encoder}
We learn representations of observations and actions  by feeding them to a recurrent network. Given the current observation $o_t$, we use pre-trained word embeddings to represent $o_t$ as a sequence of $d$-dimensional vectors $\mathbf{x}_t^1, \dots, \mathbf{x}_t^N$, where each $\mathbf{x}_t^k \in \mathbb{R}^d$ is the word embedding of the $k$-th observed token $o_t^k$, $k = 1, \dots, N$.
Then, a (bidirectional) GRU encoder \cite{cho2014gru} is used to process the sequence $\mathbf{x}_t^1, \dots, \mathbf{x}_t^N$ to get the representation of the current observation:
$\mathbf{o}_t = \mathbf{h}^N_{t}$, where $\mathbf{h}^k_{t} = GRU(\mathbf{h}^{k - 1}_{t}, \mathbf{x}_t^k)$, for $k = 1, \dots, N$.
In a similar way, given the set $A_t$ of admissible actions at time step $t$, we learn representations of each action $a \in A_t$.

\subsection{Context Encoder}
\label{subsec:context_encoder}

A key challenge for our RL agent is in modeling context, i.e. the history of observations.
We model the context using another recurrent encoder over the observation representations $\mathbf{o}_t$.
We use a GRU network to encode the sequence of previous observations up to $o_t$ into a vector $\mathbf{s}_t = GRU(\mathbf{s}_{t-1}, \mathbf{o}_t)$.
We refer to $s_t$ as the state vector, or the context encoding. The context encoding will be used in addition to the commonsense knowledge in the final action prediction.

\subsection{Dynamic Commonsense Subgraph}
\label{subsec:dynamic_commonsense-subgraph}
Our model retrieves commonsense knowledge from ConceptNet in the form of a graph. The graph $G_C^t$ is updated dynamically at each time step $t$. $G_C^t$ is constructed by mapping the textual observation $o_t$ at time $t$ to ConceptNet and combining it with the graph at previous time step $G_C^{t-1}$.
We used \textit{spaCy} (\url{https://spacy. io}) to extract noun chunks, and then performed a max sub-string match with all the concepts in ConceptNet. This results in a set of entities $e_t$ for the observation $o_t$ at time $t$.  
We then combine the concepts from  $G_C^{t-1}$ and $e_t$ to get $E_{t}$. $E_t$ consists of all the concepts observed by the agent until time step $t$, including the description of the room, the current observation, and the objects in the inventory. Given $E_t$, we describe three different techniques to automatically extract the commonsense graph $G_t$ from external knowledge.



\vspace{0.5mm}

\noindent \textbf{(1) \dc}: This is the baseline approach to construct $G_C^t$. We fetch direct links between each of the concepts in $E_t$ from ConceptNet. 

\vspace{0.5mm}

\noindent\textbf{(2) \cdc}: Since the goal of the agent is to clean up the house by putting \texttt{objects} into its appropriate \texttt{containers} such as \small \texttt{apple} $\Rightarrow$ \texttt{refrigerator}, \normalsize, we hypothesize that adding links only between objects and containers may benefit the agent instead of links between all concepts as done by \textit{Direct Connections}, as we might overwhelm the agent with noise. 
To accomplish this goal, we split the entities $E_t$ into objects and containers. Since we know the entities from the inventory in $E_t$ constitutes objects, no explicit labelling is needed as we consider the remaining entities as containers. We retain only the edges between objects and containers from ConceptNet. 

\vspace{0.5mm}

\noindent\textbf{(3) \hop}: Previous techniques only focus on connecting links between observed concepts $E_t$ from external knowledge. In addition to the direct relations, it may be beneficial to include concepts from external knowledge that is related to $E_t$ but has not been directly observed from the game. Therefore, for each concept in $E_t$, we include all its neighboring concepts and associated links.



\subsection{Knowledge Integration}
\label{subsec:knowledge_integration_encoder_coattention}

We enhance the text-based RL agent by allowing it to jointly contextualize information from both the commonsense subgraph and the observation representation. We call this step knowledge integration.
We encode the commonsense graph using a graph encoder followed by a co-attention layer. 

\vspace{1mm}

\noindent{\bf Graph encoder:} The graph $G_C^t$ is encoded 
as follows: First, we use pretrained KG embeddings (Numberbatch) to map the set of nodes $\mathcal{V}_t$ to a feature matrix $[\mathbf{e}_t^1, \dots, \mathbf{e}_t^{|\mathcal{V}_t|}] \in \mathbb{R}^{f \times |\mathcal{V}_*^t|}$. Here,
 $\mathbf{e}_t^i \in \mathbb{R}^f$ is the (averaged) embedding of words in node $i \in \mathcal{V}_*^t$. Following \cite{lu2017knowing}, we also add a \textit{sentinel} vector to allow the attention modules to not attend to any specific nodes in the subgraph. 
These node embeddings are updated at each time step by message passing between the nodes of $G_c^t$ with Graph Attention Networks (GATs) \cite{velickovic2018graph} to get $\{{\bf z}_t^1, {\bf z}_t^2 \cdots {\bf z}_t^{|\mathcal{V}_t|}\}$ using multi-head graph attention, resulting in a final graph representation that better captures the conceptual relations between the nodes in the subgraph.


\vspace{1mm}

\noindent{\bf Co-Attention:}
In order to combine the observational context and the retrieved commonsense graph, we consider a bidirectional attention flow layer between these representations to re-contextualize the graph for the current state of the game \cite{seo2016bidirectional,yu2018qanet}. 
\noindent Similar to \cite{yu2018qanet}, we compute a similarity matrix $S \in \mathbb{R}^{N \times |\mathcal{V}_C^t|}$ between the context and entities in the extracted common sense subgraph using a trilinear function. In particular, the similarity between $j^{th}$ token's context encoding $\mathbf{h}_t^j$ and $i^{th}$ node encoding ${\bf z}_t^i$ in the commonsense subgraph is computed as: {$S_{ij} = {\bf W}_0^T[{\bf z}_t^i; {\bf h}_t^j; {\bf z}_t^i \circ {\bf h}_t^j]$} \normalsize where $\circ$ denotes element-wise product, $;$ denotes concatenation and $\mathbf{W}_0$ is a learnable parameter. We use the softmax function to normalize the rows (columns) of $S$ and get the similarity function for the common-sense knowledge graph $\bar{S}_G$ (context representation $\bar{S}_O$). 
The commonsense-to-context attention is calculated as {$A= \bar{S}_G^T\cdot O$} and the context-to-common sense attention is calculated as {$B= \bar{S}_G\bar{S}_O^T\cdot G$}, where $G=[{\bf z}_t^1, {\bf z}_t^2, \cdots {\bf z}_t^{|\mathcal{V}_C^t|}]$ and $O=[{\bf h}_t^1, {\bf h}_t^2 \cdots {\bf h}_t^N]$ are the commonsense graph and observation encodings. The attention vectors are then combined together and the final graph encoding vectors $\mathbf{G}$ are calculated as {$\mathbf{W}^\top [\mathbf{G};\mathbf{A};\mathbf{G}\circ \mathbf{A};\mathbf{G}\circ \mathbf{B}]$} where $W$ is the learnable parameter.
Finally, we get the commonsense graph encoding $\mathbf{g}_i^t$ for each action $a_i \in A_t$ by applying a general attention over the nodes using the state vector and the action encoding $[\mathbf{s}_t;\mathbf{a}_i^t]$ \cite{luong-etal-2015-effective}. The attention score for each node is computed as {$\mathbf{\alpha}_i =[\mathbf{s}_t;\mathbf{a}_i^t] \mathbf{W}_g \mathbf{G}$}, and the commonsense graph encoding for action $\mathbf{a}_i^t$ is given as  $\mathbf{g}_i^t =\mathbf{\alpha}_i^\top G$. \normalsize

\subsection{Action Selection}

The action score for each action $\hat{a}_i^t$  is computed based on the context encoding $\mathbf{s}_t$, the commonsense graph encoding $\mathbf{g}_i^t$ and the action encoding $\mathbf{a}_i^t$.
We concatenate these encoding vectors into a single vector $\mathbf{r}_i^t = [\mathbf{s}_t; \mathbf{g}_i^t; \mathbf{a}_i^t]$.
Then, we compute probability score for each action $a_i \in A_t$ as 
{$\mathbf{p}_t = softmax(W_1 \cdot ReLU(W_2 \cdot \mathbf{r}_t + \mathbf{b}_2) + \mathbf{b}_1)$};
where $W_1, W_2, \mathbf{b}_1$, and $\mathbf{b}_2$ are learnable parameters of the model. The final action chosen by the agent is then given by the one with the maximum probability score, namely $\hat{a}_t = \arg \max_i p_{t,i}$.

\section{Experiments}
\label{sec:experiments}

\begin{figure*}[!h]
    \centering
        \includegraphics[trim={0.4cm 0.0cm 0.75cm 0.75cm},clip, scale=0.37]{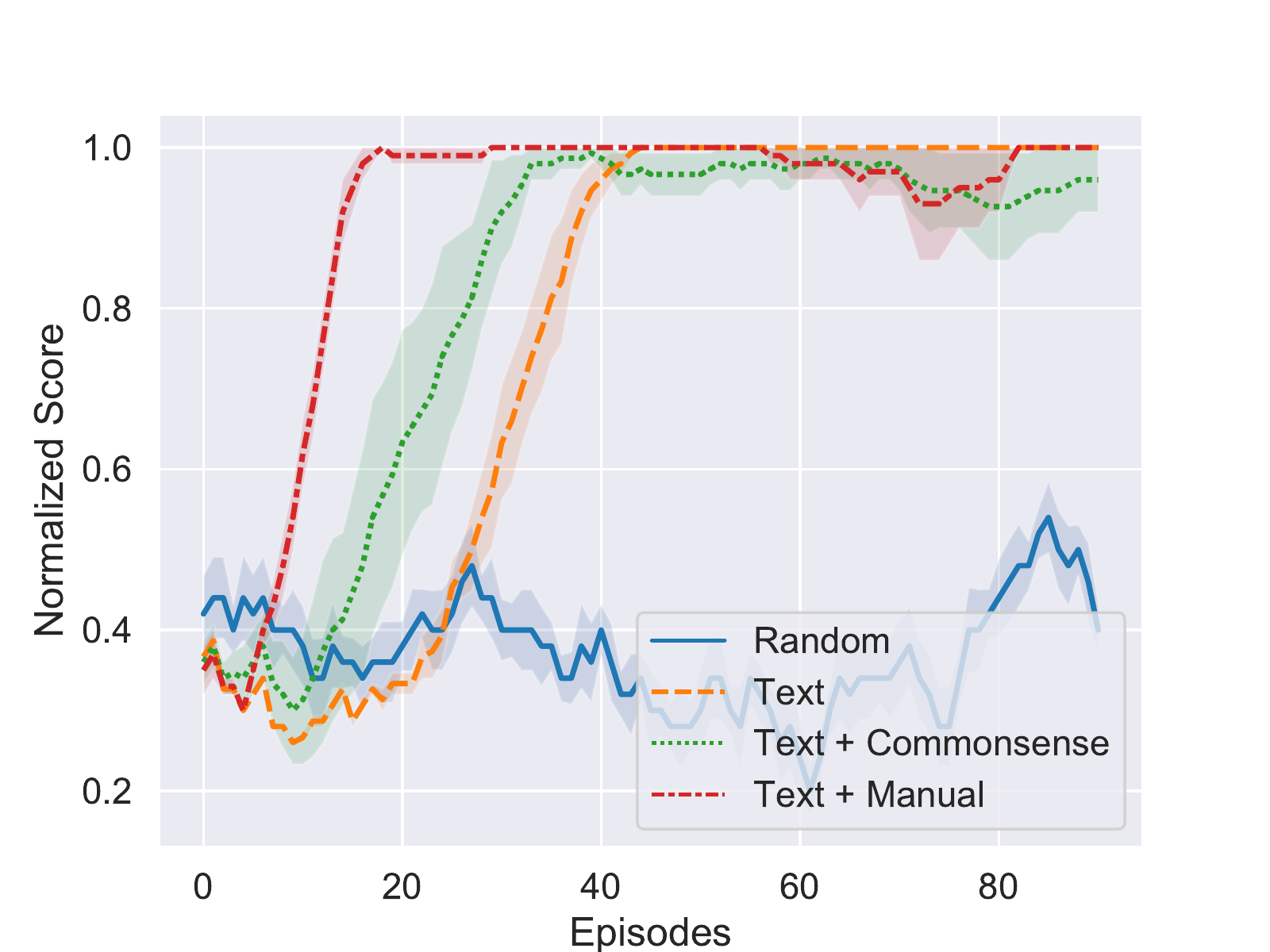}
        \includegraphics[trim={0.4cm 0.0cm 0.75cm 0.75cm},clip,scale=0.37]{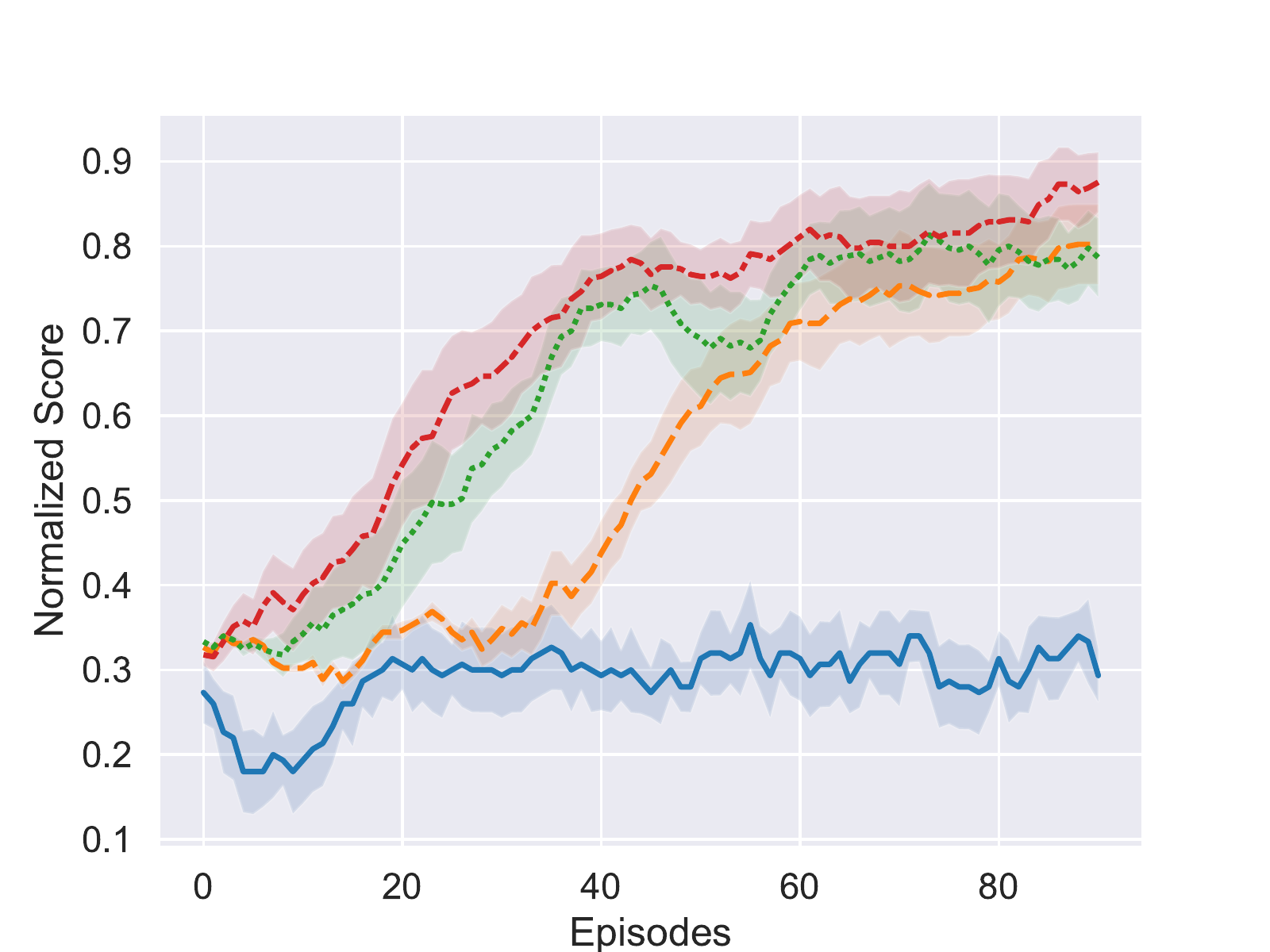}
        \includegraphics[trim={0.4cm 0.0cm 0.75cm 0.75cm},clip,scale=0.37]{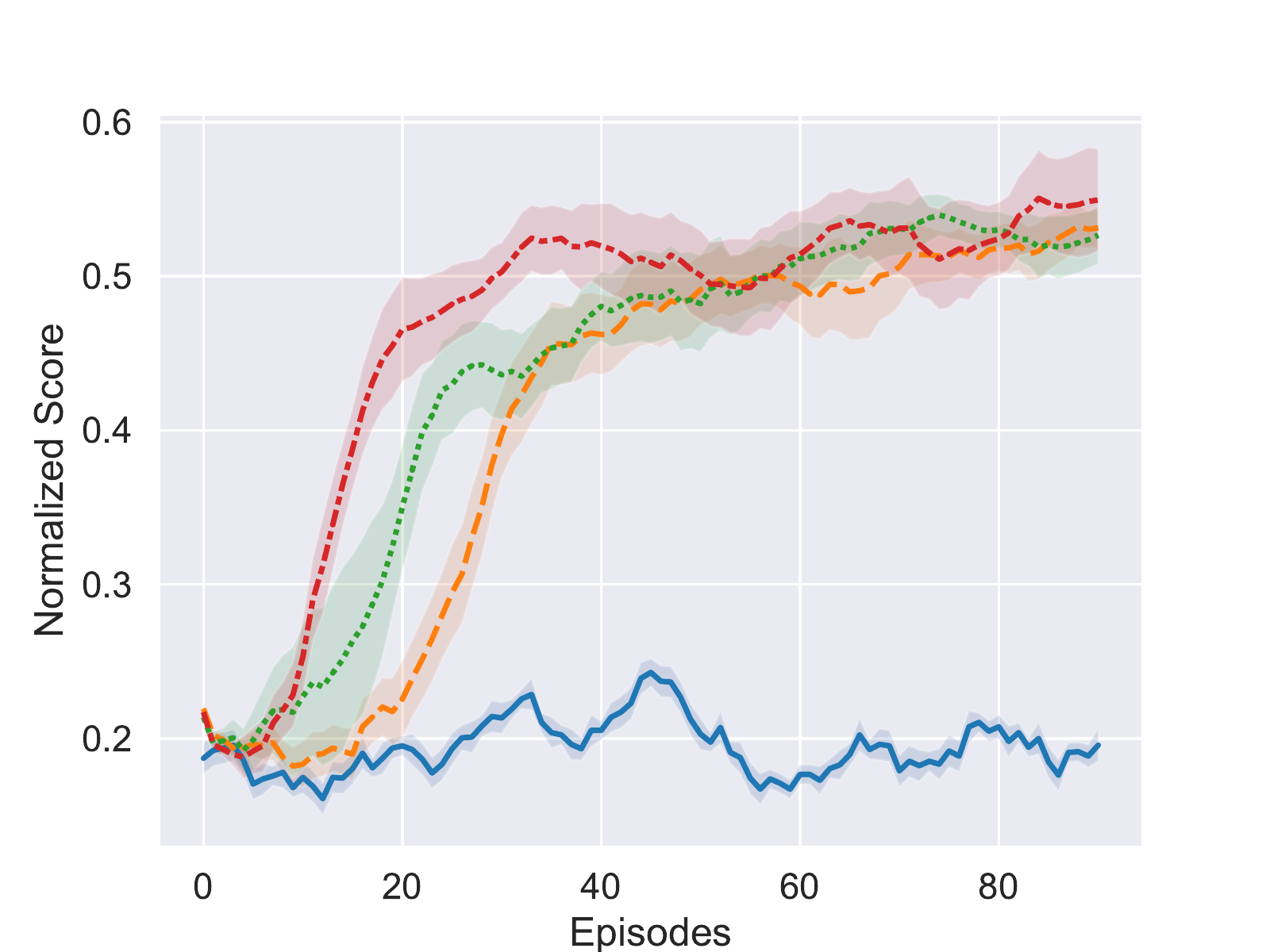}\\
        \includegraphics[trim={0.5cm 0.0cm 0.75cm 0.75cm},clip,scale=0.37]{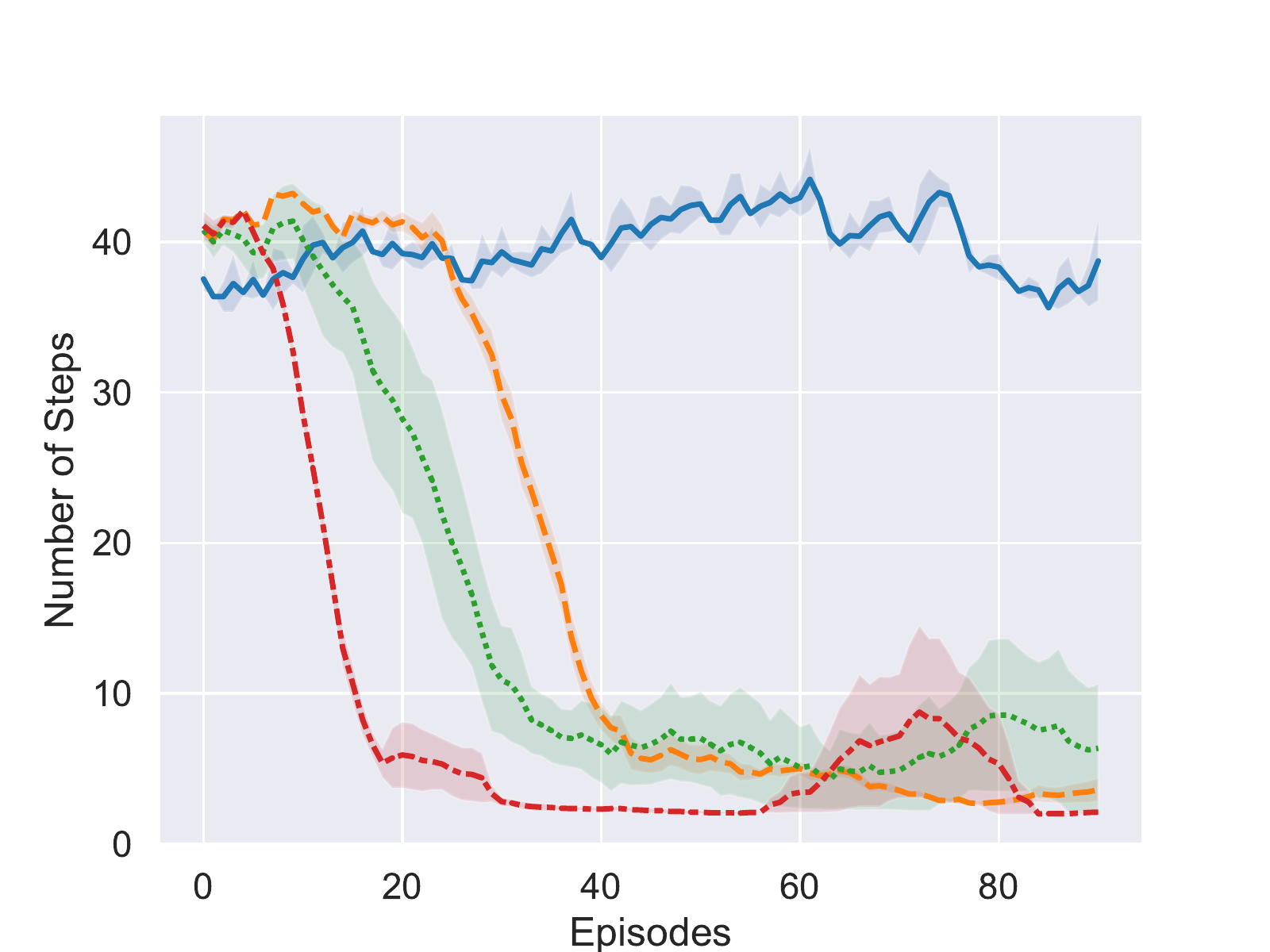}
        \includegraphics[trim={0.5cm 0.0cm 0.75cm 0.75cm},clip,scale=0.37]{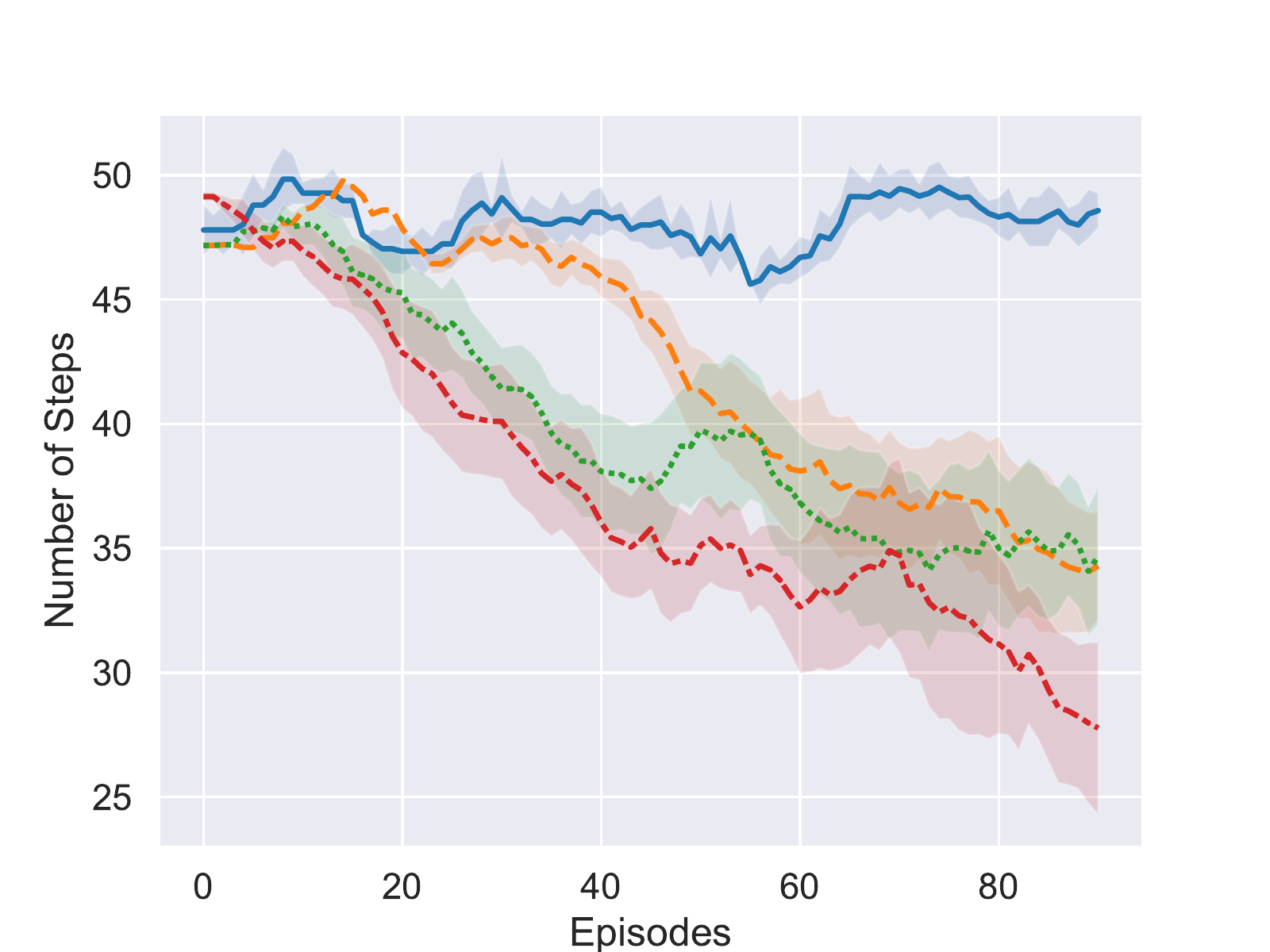}
        \includegraphics[trim={0.5cm 0.0cm 0.75cm 0.75cm},clip,scale=0.37]{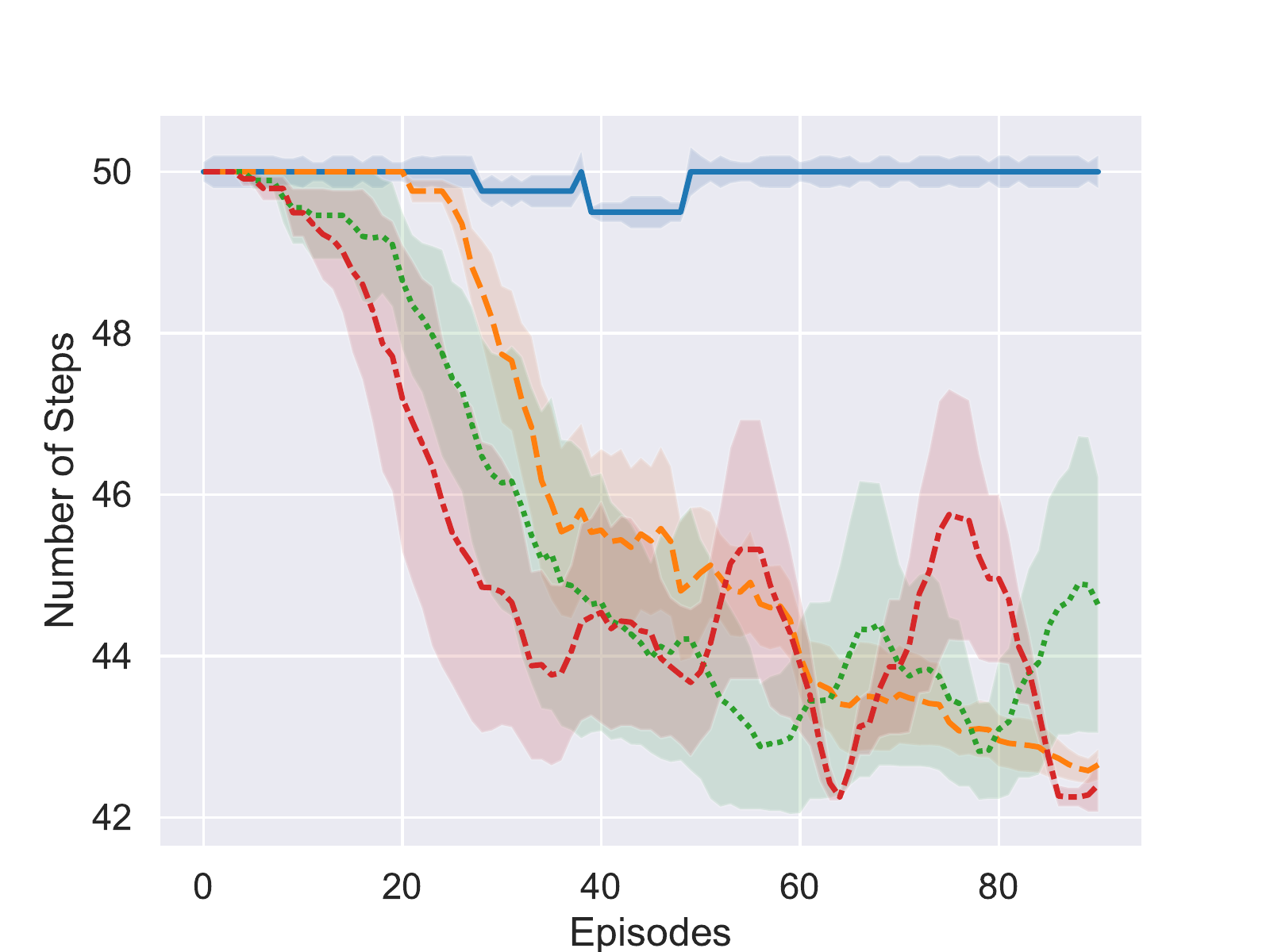}
        %
        %
    \caption{Performance evaluation (showing mean and standard deviation averaged over $10$ runs) for the three difficulty levels: Easy (left), Medium (middle), Hard (right) using normalized score and the number of steps taken.}
    \label{fig:res}
    \vspace{-4mm}
\end{figure*}

In this section, we report the results of our experiments on the \twcshort\ games. Given that the quality (correctness and completeness) of \twcshort\ has already been evaluated (c.f. Section~\ref{subsec:verifying_twc_annotations}), these experiments primarily focus on showing that: (1) agents that utilize commonsense knowledge can achieve better performance on \twcshort{} than their text-based counterparts; (2) \twcshort\ can aid research in the use of commonsense knowledge because of the gap between human performance and the commonsense knowledge agents.

\vspace{1mm}

\noindent \textbf{Experimental Setup:} We measure the performance of the various agents using: (1) the normalized score (score achieved $\div$ maximum achievable score); and (2) the number of steps taken. Each agent is trained for $100$ episodes and the results are averaged over $10$ runs. Following the winning strategy in the {\em FirstTextWorld Competition}~\cite{Adolphs2019LeDeepChefDR}, we use the Advantage Actor-Critic framework~\cite{mnih2016asynchronous} to train the agents using reward signals from the training games.

\subsection{RL Agents in \twcshort}
\label{subsec:sample_efficient_rl}
\label{subsec:rl_agents_in_twc}

We evaluate our framework on the \twcshort{} cleanup games (as described in Section \ref{sec:twc_games}).
For comparison, we consider a random agent that randomly picks an action at each time step. 
We consider two types of experiment settings based on the type of information available to the RL agents: (1) {\tt Text-based} RL agents have access to the textual description (observation) of the current state of the game provided by the \twcshort{} environment; and (2) {\tt Commonsense-based} RL agents have access to both the observation and ConceptNet. 

\vspace{1mm}

\noindent \textbf{Text-only Baseline Agents:} As baselines, we picked various SOTA text-based agents that utilize observation only: (1) \textbf{LM-NSP} uses language models such as \textit{BERT}~\cite{devlin2019bert} and \textit{GPT2}~\cite{radford2019language} with the observation and the action pair as a Next Sentence Prediction (NSP) task; (2) \textbf{LSTM-A2C} \cite{narasimhan2015language} uses the observed text to select the next action;  (3) \textbf{DRRN}~\cite{he2016deep} utilizes the relevance between the observation and action spaces for better convergence; and (4) \textbf{KG-A2C}~\cite{Ammanabrolu2020Graph} uses knowledge of the game environment generated from the observation to guide the agent's exploration. For these baselines, we use GloVe~\cite{pennington2014glove} embeddings for text.

The results on these baselines are reported in Table~\ref{tab:generalization_res}. For each difficulty level, we report: the agents' performance; the optimal number of steps to solve the game\footnote{The optimal number of steps were computed by considering the objects already in the agent's possession, the number of objects to ``put'' (goals), and the number of rooms in the instance.}; and the human performance. 
The performance of GPT2-NSP and BERT-NSP shows that even powerful pretrained models if not tuned to this task have difficulty in these commonsense RL games, as they do not capture commonsense relationships between entities.  Baselines such as LSTM-A2C, DRRN, and KG-A2C have a competitive advantage over the LM-NSP baselines, as they effectively adapt to the sequential interaction with the environment to improve performance. Among these baselines, DRRN and KG-A2C perform better than LSTM-A2C as they utilize the structure of the state and action spaces for efficient exploration of the environment.

\vspace{1mm}

\noindent \textbf{Commonsense-based agents:} We introduce commonsense knowledge in two ways. The first is (Text + Numberbatch) by replacing GloVE embeddings in the LSTM-A2C agent with Numberbatch (\textit{Nb}) embeddings~\cite{speer2017conceptnet} which were trained on text and ConceptNet. This is the naive approach to augment text information with commonsense knowledge. The results in Table~\ref{tab:generalization_res} show that introducing \textit{Nb} embeddings allows achieving a noticeable gain (an average of $3$ steps in easy and $7$ steps in medium level games) over GloVe embeddings.

\begin{figure*}[ht]
    \centering
        \includegraphics[scale=0.427]{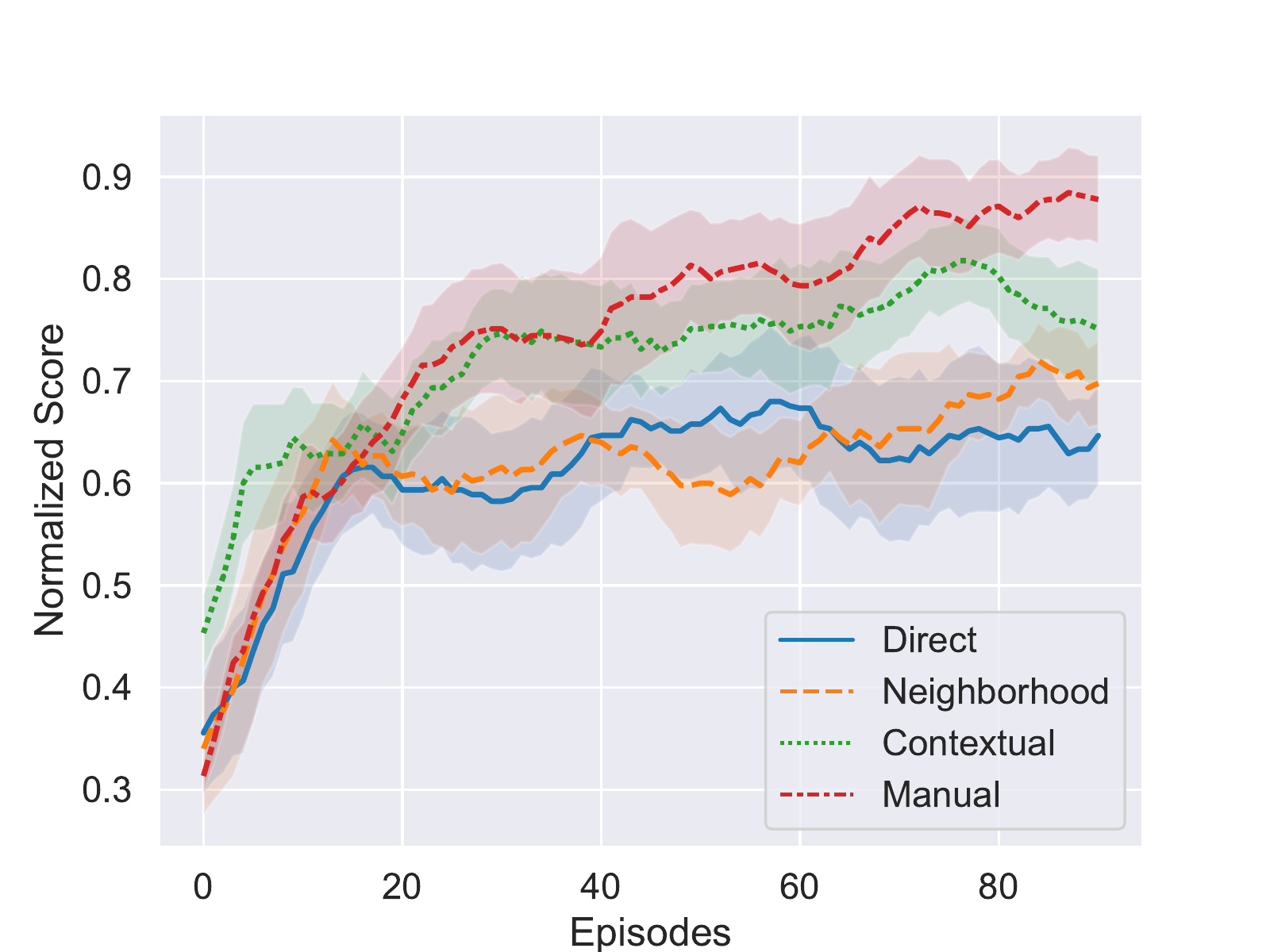}
        \includegraphics[trim={0.3cm 0.0cm 0.75cm 1.25cm},clip,scale=0.427]{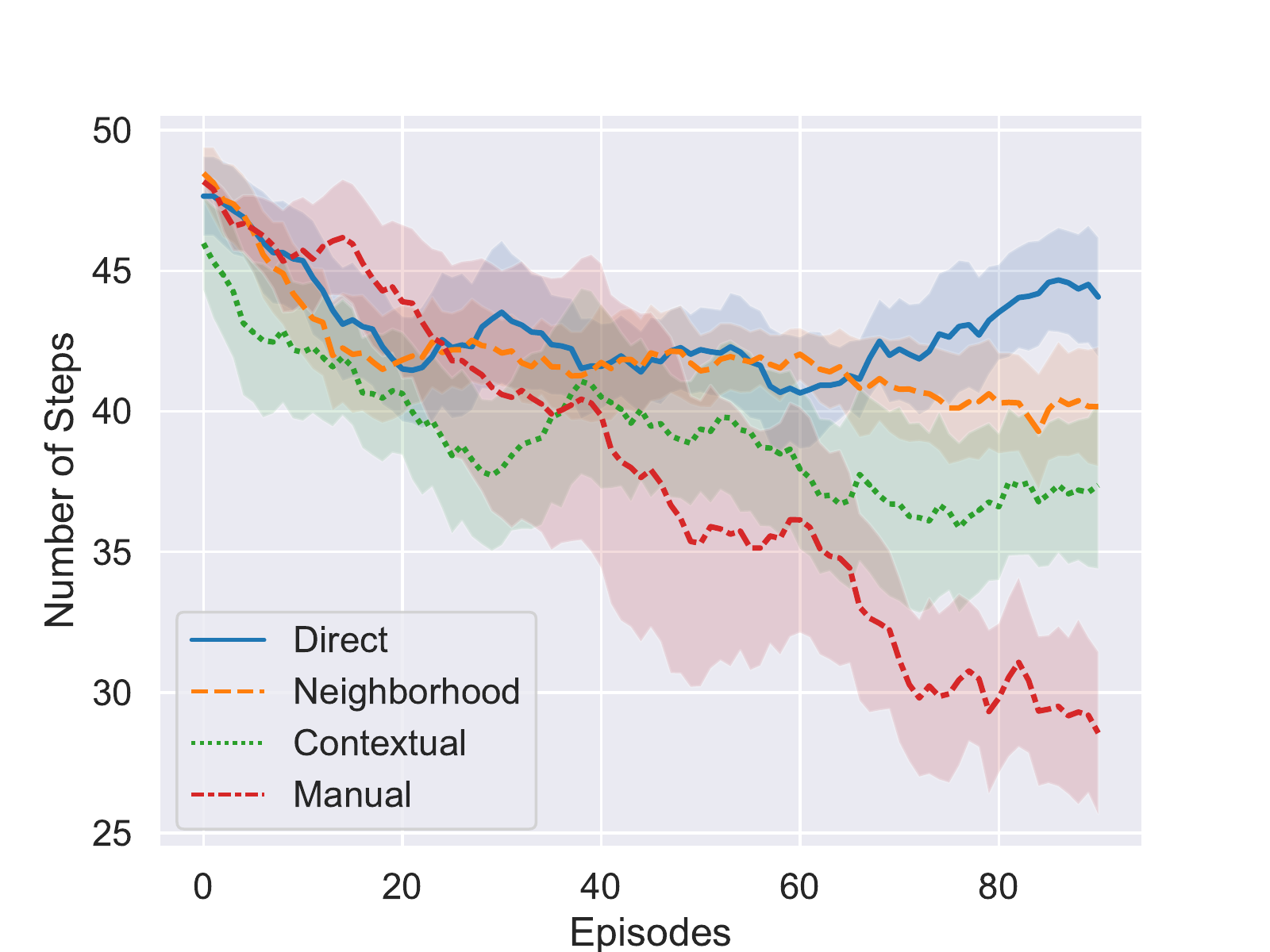}
    \caption{{Performance for the medium level (train-set) games (showing mean and standard deviation averaged over $3$ runs) with the different techniques for the commonsense sub-graph extraction.}}
    \label{fig:world_types}
    \vspace{-2mm}
        %
        %
\end{figure*}

In order to explicitly use commonsense knowledge, we experiment with the three different mechanisms outlined in Section~\ref{subsec:context_encoder} for retrieving relevant information from ConceptNet: ({\tt DC}, {\tt CDC} and {\tt NG}).
These methods retrieve both the concepts and structure in the relevant sub-graphs from ConceptNet, which are leveraged by our co-attention mechanism (Section~\ref{subsec:knowledge_integration_encoder_coattention}).  The comparison of the agents' performance with different retrieving mechanisms is shown in Fig~\ref{fig:world_types}. The results show that \texttt{CDC} performs the best among other mechanisms, particularly compared to \texttt{DC}. Unlike {\tt DC} that includes all the links between observed concepts from ConceptNet, \texttt{CDC} restricts links to those between observed objects and {\em containers}. This selection of relevant links from ConceptNet improves the performance of the agent. 


Given that $CDC$ performs best, we compare  results on text-based models with $CDC$-augmented commonsense knowledge to other baselines. Table~\ref{tab:generalization_res} shows results for text-based agents initialized with GloVe or \textit{Nb} embeddings, and augmented with commonsense knowledge. We see that the commonsense-based RL agents perform better than text-based RL agents in the easy and medium level games. This is not surprising, as these instances mostly involve picking an object and placing it in a container in the same room. Both the text-based and commonsense RL agents struggle in the hard level, as these games have more than one room and multiple objects and containers. 
We also
notice that the average number of steps taken by the commonsense-based RL agents are noticeably lower than the other agents as it efficiently uses commonsense knowledge to rule out implausible actions. This proves that \twcshort\ is a promising test-bed where commonsense knowledge helps.

Our results show that \twcshort\ still has much room for improvement in terms of retrieving and combining knowledge with observations and feedback from the environment in a sample-efficient manner. As a starting point for showing that there is headroom, we switched the retrieval mechanism to manually selected information from ConceptNet. 
We manually retrieved the relevant commonsense knowledge by extracting the commonsensical paths between entities in ConceptNet, corresponding to objects in the \twcshort{} games and their goal locations. The manual subgraph includes all the relevant shortest paths between an object and its location, within a 2-hop neighborhood expansion of both nodes. 
Since the extracted subgraph can be very large even for the easy games, further pruning was performed to remove noise. We emphasize that the manual annotation can be error-prone or result in manual subgraphs that lack potentially useful information. 
Thus, the manual graphs should not be taken as a gold standard. However, we are exploring other manual retrieval process to understand if better commonsense retrieval approaches can bring improvements in the future.
In Table~\ref{tab:generalization_res}, agents that are augmented with the manual graph perform better than the other automated retrieval mechanisms (average reduction of $2-5$ steps on easy and medium). 
Fig~\ref{fig:res} shows training curves for the Text-only, Text+Commonsense and Text+Manual agents on the three difficulty levels. We notice that infusing commonsense knowledge allows achieving faster convergence both in terms of the number of steps taken by the agents and the final score. 
We found that the extracted manual subgraphs is not perfect as can be seen in the training curves for medium and hard levels.

\vspace{1mm}

\noindent \textbf{Human Performance on \twcshort}: 
We also present the results of human performance in \twcshort\  (outlined in Section~\ref{subsec:human_performance_twc}). The \textbf{O} and \textbf{H} columns in Table~\ref{tab:generalization_res} (two per condition) present these results. A quick comparison of these numbers reveals two major results: (1) human performance \textbf{H} is very close to the optimal number of steps \textbf{O} in all 3 conditions; and (2) there is significant headroom between \textbf{H} and all of the other agents in the table, include the ones with the manual graph. This confirms that there is still much progress to be made in retrieving and encoding the commonsense knowledge effectively to solve such problems; and that \twcshort\ can spur further research.

\subsection{Generalization}
\label{subsec:results_generalization}

Table \ref{tab:generalization_res} reports the results both for test games that belong to the same distribution used at training time (\textbf{IN}), and games that were generated from a different set of entities (\textbf{OUT}). We see a similar trend on both these settings.
The commonsense-enhanced agent outperforms the text-only agent in all cases. 
However, all agents including those that utilize commonsense knowledge show similar drop in performance from {\bf IN} to {\bf OUT} distribution. This is in contrast to the use of the knowledge graphs in other NLP tasks such as textual entailment where knowledge graphs have shown to be robust to changes in the underlying (training and testing) environment~\cite{kapanipathi2020infusing,chen2018}. The task of designing knowledge-enabled agents that are robust to such changes is another open challenge for the community that can be evaluated by \twcshort.

\noindent \textbf{Results Summary: } Our results establish that  \twcshort~is an environment where agents augmented with commonsense knowledge show better performance than their text-based counterparts. Based on the experiments with manually retrieved sub-graphs, optimal steps, and the human performance numbers, we show that~\twcshort\ has enough headroom for future research efforts to: (1) retrieve more relevant commonsense knowledge for KBs; and (2) for new agents/techniques to exploit such knowledge.


\begin{table*}[ht]
\centering
\resizebox{1.0\textwidth}{!}{%
\begin{tabular}{ll|cccc|cccc|cccc}
 &  & \multicolumn{4}{c|}{\textbf{Easy}}& \multicolumn{4}{c|}{\textbf{Medium}} & \multicolumn{4}{c}{\textbf{Hard}} \\ \cline{3-14} 
 &  & \textbf{O} & \textbf{H} & \textbf{\#Steps} & \textbf{Norm. Score} & \textbf{O} & \textbf{H} &\textbf{\#Steps} & \textbf{Norm. Score} & \textbf{O} & \textbf{H} & \textbf{\#Steps} & \textbf{Norm. Score} \\ 
 \hline
 
 
\multicolumn{1}{l|}{\multirow{11}{*}{\rotatebox[origin=c]{90}{{\bf IN}}}} & \textbf{GPT2-NSP} & \multirow{5}{*}{{\rotatebox[origin=c]{90}{2.00 $\pm$ 0.00}}} & \multirow{5}{*}{{\rotatebox[origin=c]{90}{2.12 $\pm$ 0.49}}} & 30.36   $\pm$ 0.00 & 0.64 $\pm$ 0.00 & \multirow{5}{*}{\rotatebox[origin=c]{90}{3.60 $\pm$ 0.55}} &\multirow{5}{*}{\rotatebox[origin=c]{90}{5.33 $\pm$ 2.06}} & 42.12 $\pm$ 0.00 & 0.70 $\pm$ 0.00 & 
\multirow{5}{*}{\rotatebox[origin=c]{90}{15.00 $\pm$ 2.00}} &\multirow{5}{*}{\rotatebox[origin=c]{90}{15.00 $\pm$ 3.29}} & 50.00 $\pm$ 0.00 & 0.36 $\pm$ 0.00 \\

\multicolumn{1}{l|}{} & \textbf{BERT-NSP} &  & 
& 25.20 $\pm$ 0.00 & 0.76 $\pm$ 0.00 & & 
& {34.72} $\pm$ {0.00} & {0.88} $\pm$ {0.00} & & 
& {50.00} $\pm$ {0.00} & 0.52 $\pm$ 0.00 \\
\multicolumn{1}{l|}{} & \textbf{LSTM-A2C} &  &  
& 17.59  $\pm$ 3.11 & 0.86 $\pm$ 0.04 &  &  
& 37.99 $\pm$ 6.03 & 0.74 $\pm$ 0.11 &  &  
& 49.21 $\pm$ 0.58 & 0.54 $\pm$ 0.04 \\
\multicolumn{1}{l|}{} & \textbf{DRRN} &  &. 
& 18.88   $\pm$ 2.69 & 0.81 $\pm$ 0.08 &  &  
& 33.41 $\pm$ 2.81 & 0.73 $\pm$ 0.06 &  &  
& 46.20 $\pm$ 4.86 & 0.44 $\pm$ 0.01 \\
\multicolumn{1}{l|}{} & \textbf{KG-A2C} &  &  
& 17.65   $\pm$ 3.62 & 0.85 $\pm$ 0.07 &  &  
& 37.18 $\pm$ 4.86 & 0.72 $\pm$ 0.07 &  &  
& 49.36 $\pm$ 7.50 & 0.46 $\pm$ 0.10\\
\cline{2-14}
\multicolumn{1}{l|}{} & \textbf{Text} &  & & & &  &  & &  &  & &  & \\
\multicolumn{1}{l|}{} & \textit{~+Commonsense} &  &  
& {14.18 $\pm$ 6.47} & {0.89} $\pm$ {0.10} &   &  
& 34.67 $\pm$ 6.65 & 0.78 $\pm$ 0.07 &   &  
& 48.45 $\pm$ 2.50 & 0.51 $\pm$ 0.10 \\
\multicolumn{1}{l|}{} & \textit{~+Manual} &  &  
& {13.70 $\pm$ 1.85} & {0.92} $\pm$ {0.03} &  &  
& 29.26 $\pm$ 0.94 & 0.88 $\pm$ 0.03 &  &  
& 46.43 $\pm$ 3.67 & 0.54 $\pm$ 0.04 \\
\multicolumn{1}{l|}{} & \textit{~+Numberbatch} &  & 
& 11.79   $\pm$ 3.04 & 0.96 $\pm$ 0.03 &  & 
& 27.10 $\pm$ 5.06 & 0.85 $\pm$ 0.06 &  &  
& 44.22 $\pm$ 4.86 & 0.57 $\pm$ 0.00 \\
\multicolumn{1}{l|}{} & \textit{~+Nb+Commonsense} &  &  
& {14.43 $\pm$ 3.08} & {0.93} $\pm$ {0.06} &   &  
& 25.11 $\pm$ 2.33 & 0.87 $\pm$ 0.04 &   &  
& 43.27 $\pm$ 0.70 & 0.45 $\pm$ 0.00 \\
\multicolumn{1}{l|}{} & \textit{~+Nb+Manual} &  &  
& 13.37 $\pm$ 5.63 & 0.92 $\pm$ 0.07 &  &  
& 23.51 $\pm$ 1.28 & 0.91 $\pm$ 0.06 &  &  
& 42.87 $\pm$ 0.65 & 0.52 $\pm$ 0.01 \\
\hline\hline



\multicolumn{1}{l|}{\multirow{11}{*}{\rotatebox[origin=c]{90}{{\bf OUT}}}} & \textbf{GPT2-NSP} & \multirow{5}{*}{\rotatebox[origin=c]{90}{2.00 $\pm$ 0.00}} 
&\multirow{5}{*}{\rotatebox[origin=c]{90}{2.24 $\pm$ 0.75}} & 40.28 $\pm$ 0.00 & 0.46 $\pm$ 0.00 & 
\multirow{5}{*}{\rotatebox[origin=c]{90}{4.40 $\pm$ 1.14}} 
&\multirow{5}{*}{\rotatebox[origin=c]{90}{4.40 $\pm$ 1.85}} & 44.96 $\pm$ 0.00 & 0.38 $\pm$ 0.00 & 
\multirow{5}{*}{\rotatebox[origin=c]{90}{14.60 $\pm$ 2.67}} 
&\multirow{5}{*}{\rotatebox[origin=c]{90}{17.67 $\pm$ 3.31}} & 50.00 $\pm$ 0.00 & 0.14 $\pm$ 0.00 \\

\multicolumn{1}{l|}{} & \textbf{BERT-NSP} &  &  & 24.76 $\pm$ 0.00 & 0.72 $\pm$ 0.00 &  &  & {41.12} $\pm$ {0.00} & {0.55} $\pm$ {0.00} &  &  & 50.00 $\pm$ 0.00 & 0.27 $\pm$ 0.00 \\

\multicolumn{1}{l|}{} & \textbf{LSTM-A2C} &  &  
& 19.89   $\pm$ 1.86 & 0.79 $\pm$ 0.01 &  &  
& 43.70 $\pm$ 5.52 & 0.52 $\pm$ 0.18 &  &  
& 50.00 $\pm$ 0.00 & 0.27 $\pm$ 0.01 \\
\multicolumn{1}{l|}{} & \textbf{DRRN} &  &. 
& 19.49   $\pm$ 4.89 & 0.84 $\pm$ 0.08 &  &  
& 40.49 $\pm$ 4.41 & 0.56 $\pm$ 0.07 &  &  
& 50.00 $\pm$ 0.00 & 0.18 $\pm$ 0.10 \\
\multicolumn{1}{l|}{} & \textbf{KG-A2C} &  &  
& 18.00   $\pm$ 3.24 & 0.87 $\pm$ 0.05 &  &  
& 43.08 $\pm$ 4.13 & 0.54 $\pm$ 0.17 &  &  
& 49.96 $\pm$ 0.00 & 0.22 $\pm$ 0.00 \\
\cline{2-14}
\multicolumn{1}{l|}{} & \textbf{Text} &  & & & &  &  & &  &  & &  & \\
\multicolumn{1}{l|}{} & \textit{~+Commonsense} &  &  
& {19.14} $\pm$ {3.32} & {0.83} $\pm$ {0.07} &  &  
& 41.01 $\pm$ 6.97 & 0.56 $\pm$ 0.13 &  &  
& 49.99 $\pm$ 0.01 & 0.28 $\pm$ 0.05 \\
\multicolumn{1}{l|}{} & \textit{~+Manual} &  &  
& {16.86 $\pm$ 2.26} & {0.89} $\pm$ {0.04} &  &  
& 39.95 $\pm$ 2.46 & 0.71 $\pm$ 0.06 &  &  
& 49.97 $\pm$ 0.04 & 0.26 $\pm$ 0.11 \\
\multicolumn{1}{l|}{} & \textit{~+Numberbatch} &  & 
& 19.77   $\pm$ 2.50 & 0.81 $\pm$ 0.15 &  & 
& 34.54 $\pm$ 2.89 & 0.80 $\pm$ 0.04 &  &  
& 49.95 $\pm$ 0.08 & 0.29 $\pm$ 0.02 \\
\multicolumn{1}{l|}{} & \textit{~+Nb+Commonsense} &  &  
& {20.84} $\pm$ {1.13} & {0.83} $\pm$ {0.03} &  &  
& 33.43 $\pm$ 2.11 & 0.71 $\pm$ 0.09 &  &  
& 50.00 $\pm$ 0.00 & 0.25 $\pm$ 0.01 \\
\multicolumn{1}{l|}{} & \textit{~+Nb+Manual} &  &  
& 18.24 $\pm$ 4.63 & 0.83 $\pm$ 0.09 &  &  
& 30.12 $\pm$ 4.62 & 0.84 $\pm$ 0.03 &  &  
& 49.99 $\pm$ 0.02 & 0.22 $\pm$ 0.05 \\
\end{tabular}%
}
\caption{Generalization results for within distribution (IN) and out-of-distribution (OUT) games}. \textbf{O} represents optimal \# steps needed to accomplish the goals. \textbf{H} represents human level performance. \textcolor{black}{All agents were restricted to a maximum of $50$ steps.}
\label{tab:generalization_res}
\vspace{-5mm}
\end{table*}

\section{Related Work}
\label{sec:related_work}

\noindent {\bf RL Environments and TextWorld}: 
Games are a rich domain for studying grounded language and how information from text can be utilized in control.
Recent work has explored text-based RL games to learn strategies
for \textit{CivII} \citep{branavan2012learning}, multi-user dungeon games \citep{narasimhan2015language}, etc.
Our work builds on the \texttt{TextWorld}~\cite{cote18textworld} sandbox learning environment.
A recent line of work on TextWorld learns symbolic representations of the agent's belief. Notably, \citet{ammanabrolu2019playing} proposed \textit{KG-DQN} and \citet{adhikari2020learning} proposed \textit{GATA}. Both approaches represent the game state as a belief graph. 
This graph is used to prune the action space, enabling efficient exploration, in a different way from our work which uses common sense. 

\vspace{1mm}
\noindent {\bf External Knowledge for Efficient RL:}
There have been few attempts on adding prior or external knowledge to RL approaches. Notably, 
\citet{garnelo2016towards} proposed \textit{Deep Symbolic RL}, which combines aspects of symbolic AI with neural networks and RL as a way to introduce commonsense priors. 
There has also been work on \textit{policy transfer}~\cite{bianchi2015transferring}, which studies how
knowledge acquired in one environment can be re-used in another one;
and \textit{experience replay}~\cite{wang2016sample,lin1992self,lin1993reinforcement} which studies how an agent's previous experiences can be stored and then later reused.
In this paper, we use commonsense knowledge as a way to improve sample efficiency in text-based RL agents. 
To the best of our knowledge, there is no prior work that {\em practically} explores how commonsense can be used to make RL agents more efficient.
The most relevant prior work is by \citet{martin:18}, who use commonsense rules to build agents that can play tabletop role-playing games. However, unlike our work, the commonsense rules in this work are manually engineered. 

\vspace{1mm}
\noindent {\bf Leveraging Commonsense:}
Recently, there has been a lot of work in NLP to utilize commonsense for QA, NLI, etc. \cite{sap2019atomic,talmor2018commonsenseqa}.
Many of these approaches seek to effectively utilize ConceptNet by reducing the noise retrieved from it~\cite{lin2019kagnet,kapanipathi2020infusing}. This is also a key challenge in \twcshort.

\section{Conclusion}

We created a novel environment (\twcshort) to evaluate the performance on RL agents on text-based games requiring commonsense knowledge.
We introduced a framework of agents which tracks the state of the world; uses the sequential context to dynamically retrieve relevant commonsense knowledge from a knowledge graph; and learns to combine the two different modalities. 
Our agents equipped with common sense achieve their goals with greater efficiency and less exploration when compared to a text-only model, thus showing the value of our new environments and models. 
Therefore, we believe that our \twcshort{} environment provides interesting challenges and can be effectively used to fuel further research in this area.

\vspace{1mm}

\section*{Reproducibility}

\noindent  To ensure the wide and unrestricted usage of the \twcshort\ environment, we release the \twcshort\ environment (with \textit{anonymized} human annotations), code to generate text-based games and the sample agents used in this paper here: \url{https://github.com/IBM/commonsense-rl}.


\appendix
\section{Overlap between \twcshort{} and ConceptNet}
\label{sec:overlap}
There is definitely some overlap between the resources used to build \twcshort{} and ConceptNet.
However, as we discuss below, the overlap is limited and it is non-trivial for the agents to explore the knowledge graph and retrieve relevant commonsense knowledge. Indeed, only 12.2\% of the \textit{goal} entity-location pairs defined in the \twcshort{} dataset can be directly matched to a single triplet in ConceptNet. Hence, we can state that it is fair and challenging to use ConceptNet as an external source of information. At the same time, we claim that external commonsense knowledge sources can be actually useful in solving text-based games. We submit that 85.9\% of the unique entities in TWC match exactly one node in ConceptNet. Moreover, 66.1\% of the time, the goal location of a given entity is in its 3-hop neighborhood in ConceptNet (42.7\% for a 2-hop neighborhood). This shows that an external source of commonsense like ConceptNet can help to reduce exploration while solving the games, but needs to be explored effectively.
As an example, with reference to Figure \ref{fig:graph_visualization}, the relation between the entity \texttt{cap} and the goal location \texttt{hat\_rack} can be derived from ConceptNet by following the path: \texttt{cap} $\rightarrow$ \texttt{relatedTo} $\rightarrow$ \texttt{head} $\rightarrow$ \texttt{relatedTo} $\rightarrow$ \texttt{hat} $\rightarrow$ \texttt{atLocation} $\rightarrow$ \texttt{hat\_rack}.

\begin{figure*}[!htb]
    \centering
    \includegraphics[width=\textwidth]{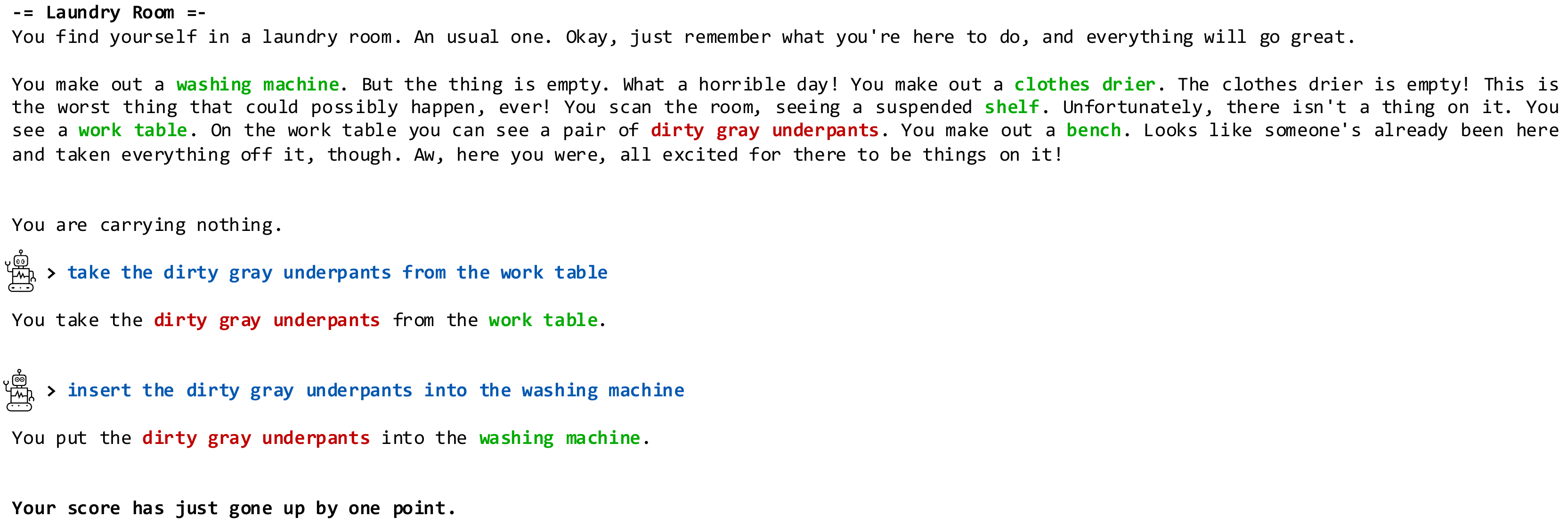}
    \caption{Example of a game walkthrough belonging to the \emph{easy} difficulty level. Best viewed in colors. Highlights are not available to the agents and are shown for illustrative purpose only.}
    \label{fig:easy_game_walkthrough}
\end{figure*}

\section{Sample \twcshort{} games}

\begin{figure*}[!htb]
    \centering
    \includegraphics[width=\textwidth]{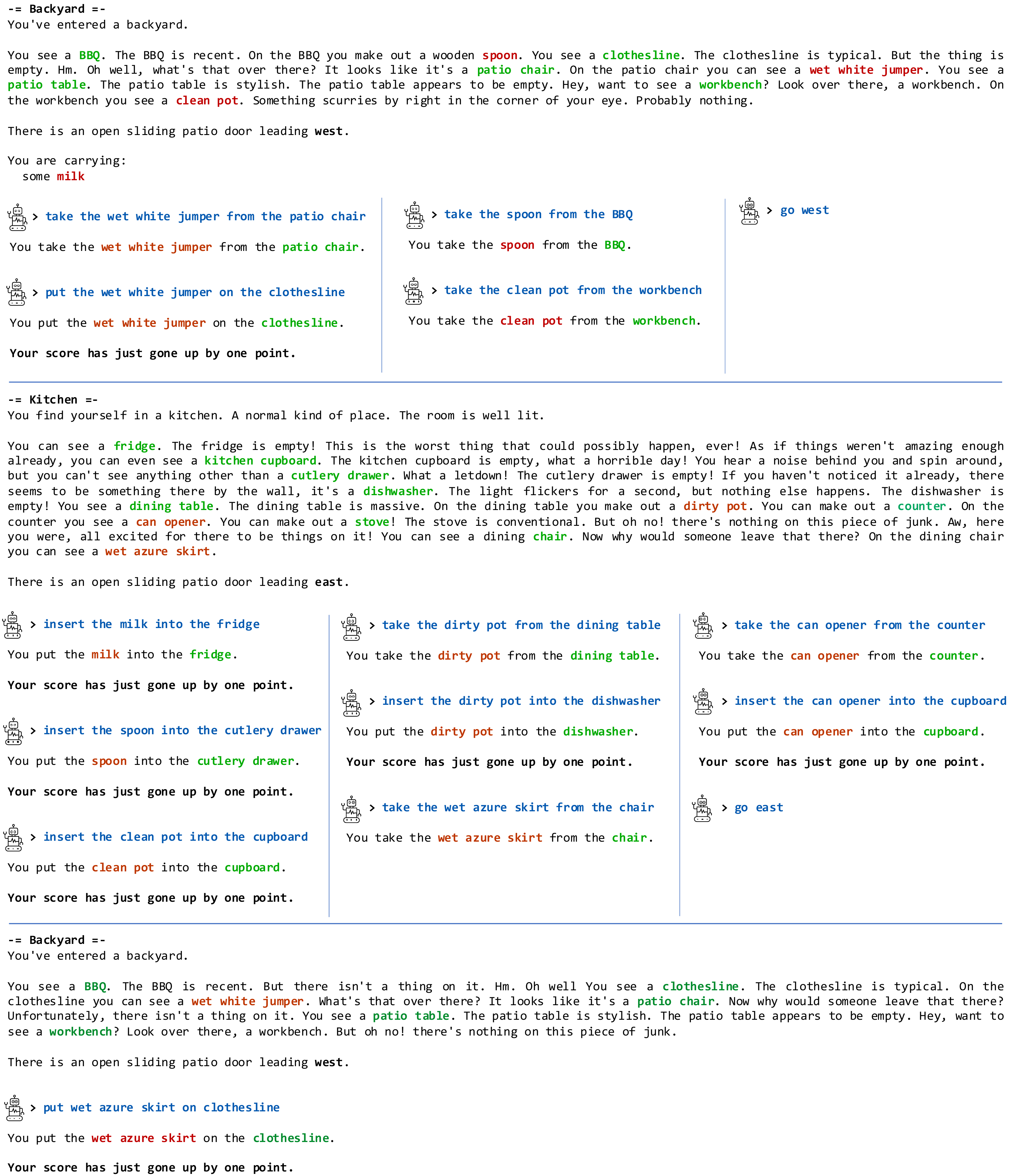}
    \caption{Example of a game walkthrough belonging to the \emph{hard} difficulty level. Best viewed in colors. Highlights are not available to the agents and are shown for illustrative purpose only.}
    \label{fig:hard_game_walkthrough}
\end{figure*}

In this section, we show and analyze an example of a \twcshort{} game instance from each difficulty level. Figures \ref{fig:easy_game_walkthrough}, \ref{fig:medium_game_walkthrough}, and \ref{fig:hard_game_walkthrough} provide such examples together with the optimal solution to each of the analyzed games. In all figures, we highlight all objects (in red), their candidate locations (in green) and the actions taken by the agent (in blue). Note that this information is not available to the agent and is only used for illustrative purposes.

Figure \ref{fig:easy_game_walkthrough} shows a walk-through of an \textit{easy} game. This game has only 1 room and 1 object. We recall that this holds for all the games in the \emph{easy} difficulty level. The game takes place in the \texttt{Laundry Room}, and the goal of the agent is to identify the correct location for the only object, in this case the \texttt{dirty gray underpants}. As all the \emph{easy} games, the agent can reach the goal with a sequence of steps consisting of only two actions. The first action is used by the agent to take the object, and then the second action is aimed at putting the object in its goal location. In general, the goal location is not unique, but in the example shown in Figure \ref{fig:easy_game_walkthrough}, there is only one correct location, namely the \texttt{washing machine}. Commonsense knowledge is required in the second step in order to detect the correct location among all the possible candidates.

The relatively large number of possible locations makes the \textit{easy} games more challenging than they might look like. In our example, the locations that are not considered commonsensical for the \texttt{dirty gray underpants} are the following: \texttt{clothes drier}, \texttt{shelf}, \texttt{work table} and \texttt{bench}. Note that the \texttt{clothes drier} could have been a commonsensical location for the entity \emph{gray underpants}, but the attribute \texttt{dirty} plays a key role. This shows that incorporating only knowledge in the form of single facts extracted from the knowledge graph is not sufficient to solve the games. On the contrary, the agent needs to aggregate commonsense knowledge from multiple triples in the knowledge graph, as previously discussed in Section \ref{sec:overlap}.

Figure \ref{fig:medium_game_walkthrough} shows an example of a more complex game belonging to the \emph{medium} difficulty level. This game is the same that the graphs in Figure \ref{fig:graph_visualization} refer to.
All \emph{medium}-level games have only 1 room and either 2 or 3 objects.
The game shown in Figure \ref{fig:medium_game_walkthrough} has 3 objects, namely a pair of \texttt{climbing shoes}, a \texttt{brown cap} and a \texttt{white cap}.
The goal locations for these objects are shown in Figure \ref{fig:visual1} and need to be selected from a pool of 5 candidate locations. However, please notice that candidate locations and objects are not provided explicitly to the agent and need to be extracted from the natural language observations. A total of $6$ steps is required to solve the game in the optimal case. These actions are reported in Figure \ref{fig:medium_game_walkthrough}. We can see that, similarly to what we have seen for the \emph{easy} games, 2 steps for each object are needed. Every time that an object is placed in its goal location, the agent receives a reward, but no reward is given for the action of taking an object.
Hence, the maximum final score that the agent can achieve is always equal to the number of objects, in this case 3.

Finally, Figure \ref{fig:hard_game_walkthrough} shows an example of the most complex games in \twcshort{}, namely the \emph{hard} games. The game includes two rooms (\texttt{Kitchen} and \texttt{Backyard}) and the agent needs to place  a total of 7 objects in the corresponding goal location. At the beginning of the game, the agent is already carrying an object (some \texttt{milk}), so it only needs to find 6 of the remaining objects. Since the game has more than 1 room, reaching the final goal may require more than 2 steps for each object. This happens because some objects may need to be carried across rooms and the in this case the agent has to visit back the initial room. In this examples, the \emph{wet azure skirt} in the \texttt{Kitchen} has to be carried back to the the \texttt{Backyard} and placed in the \texttt{clothesline}. The provided optimal solution to the analysed game consists of the 15 actions reported in Figure \ref{fig:hard_game_walkthrough}.

\section{Visualizing Attention Over the Commonsense Subgraph}
\label{sec:visual}

\begin{figure*}[h]
    \begin{subfigure}{0.39\textwidth}
    \centering
        \includegraphics[width=0.91\linewidth]{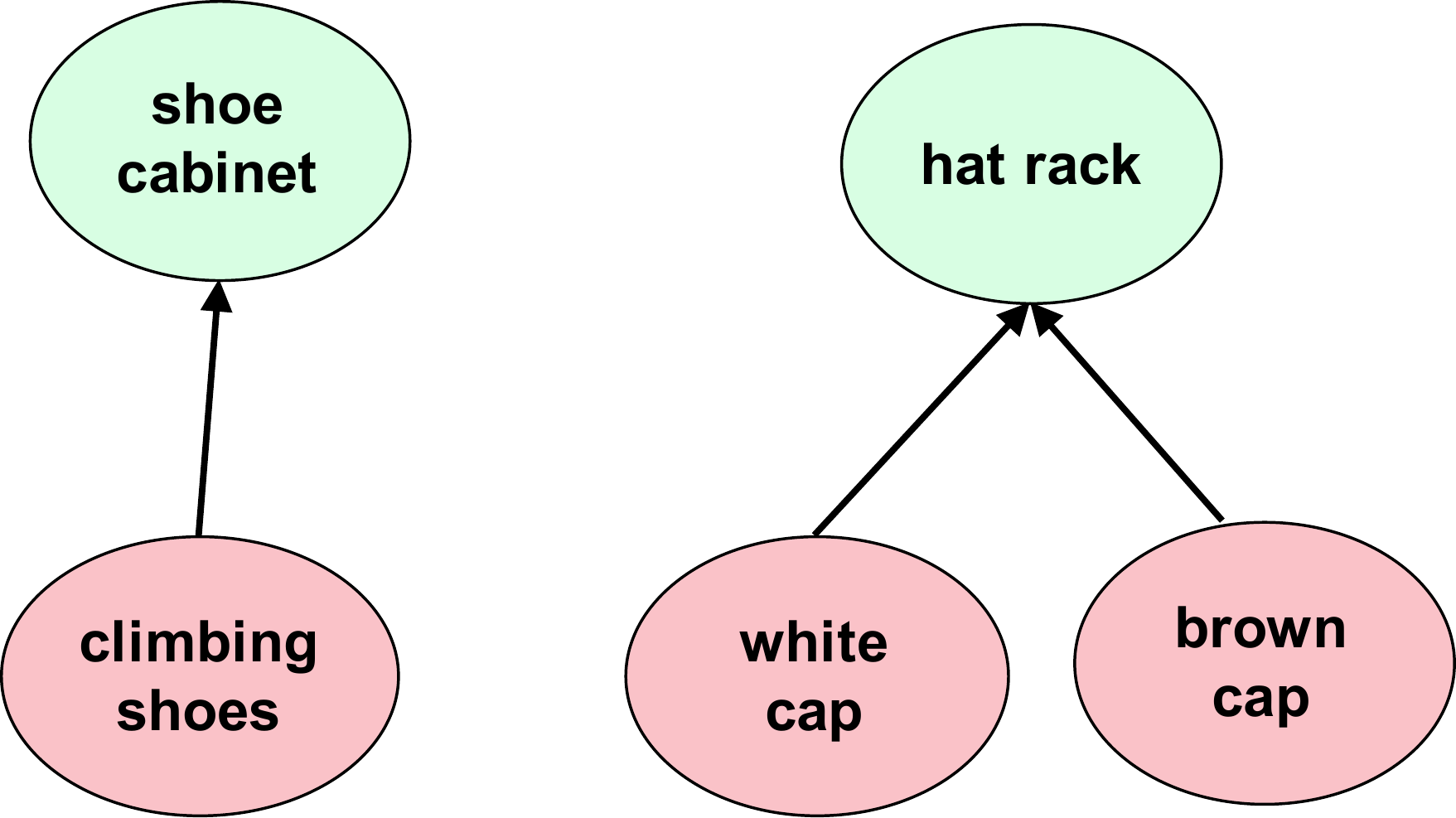}
    \caption{}
    \label{fig:visual1}
    \end{subfigure}
    \begin{subfigure}{0.6\textwidth}
    \centering
        \includegraphics[width=0.88\linewidth]{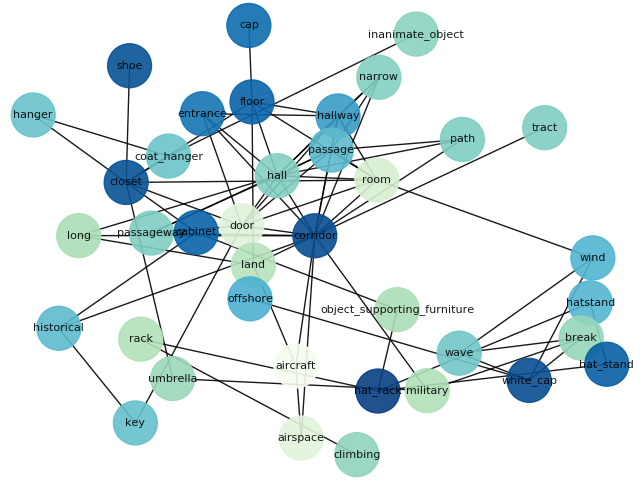}
    \caption{}
    \label{fig:visual2}
    \end{subfigure}
    \caption{A graph describing the goal state of a \twcshort{} game \textit{(a)} and a visualization of the attention weights over the ConceptNet subgraph extracted at a specific timestep of the same game instance \textit{(b)}}
    \label{fig:graph_visualization}
\end{figure*}

Our approach integrates commonsense knowledge dynamically with the context information through an attention mechanism. We can visualize the attention weights (from the general attention in the last layer) by the \emph{Text + Commonsense} agent on the commonsense graph to show how the agent is using the commonsense knowledge while interacting with \twcshort{} games.
We show the visuals for a specific game where the goal of the agent is to put the \textit{climbing shoes} in the \textit{shoe cabinet}, and the \textit{brown and white caps} in the \textit{hat rack} (see Figure \ref{fig:visual1}). The full game corresponding to this visualization was outlined in Figure \ref{fig:medium_game_walkthrough}. The attention maps for this instance are shown in Figure~\ref{fig:visual2}. We observe that the model pays attention to relevant concepts such as ``shoe'', ``hat'', ``hat rack'', ``hat stand'', etc. This figure thus offers a qualitative illustration of how commonsense knowledge is used by the \emph{Text + Commonsense} agent.

\section{Kitchen Cleanup Task: Full vs Evolve Commonsense Subgraph}
\label{sec:kitchen_ckeanup_task}
Unlike in the previous works \cite{adhikari2020learning,Ammanabrolu2020Graph}, we do not assume that the entities used in the games are known beforehand. This poses an interesting question: does the commonsense-based agents would benefit from (1) having access to the \textit{full} commonsense subgraph extracted at the beginning of the game using the aforementioned entities list or building the extracted commonsense subgraph sequentially (\textit{evolve}) based on the entities seen so far in the game. In this section, we demonstrate a very simple experiment to show why we choose \textit{evolve} setting for extracting the relevant commonsense subgraph.

We used the TextWorld~\cite{cote18textworld} environment to generate a specific game instance similar to those in \twcshort\ but used information directly from ConceptNet, and has been manually generated by an expert. We call it {\em Kitchen Cleanup}. We generate the game with $10$ objects relevant to the game, and $5$ distractor objects spread across the room. As before, the goal of the agent is to tidy the room (kitchen) by putting the objects in the right place. We create a set of realistic kitchen cleanup goals for the agent: for instance, \textit{take apple from the table} and \textit{put apple inside the refrigerator}.  Since information on concepts that map to the objects in the room is explicitly provided in ConceptNet (\texttt{Apple}  $\rightarrow$ \texttt{AtLocation}  $\rightarrow$ \texttt{Refrigerator}), the main hypothesis underlying the creation of this game is that leveraging the commonsense knowledge will let the agent achieve a higher reward while reducing the number of interactions with the environment.

The agent is presented with the textual description of a kitchen, consisting of the location of different objects in the kitchen and their spatial relationship to the other objects. The agent uses this information to select the next action to perform in the environment. Whenever the agent takes an object and puts it in the target location, it receives a reward and its total score goes up by one point. The maximum score that can be achieved by the agent in this kitchen cleanup task is equal to $10$.  In addition to the textual description, we use the commonsense knowledge subgraph extracted from ConceptNet relevant to the text description (Full vs Evolve setting). 

\begin{figure*}[t]
    \centering
    \includegraphics[width=0.44\linewidth]{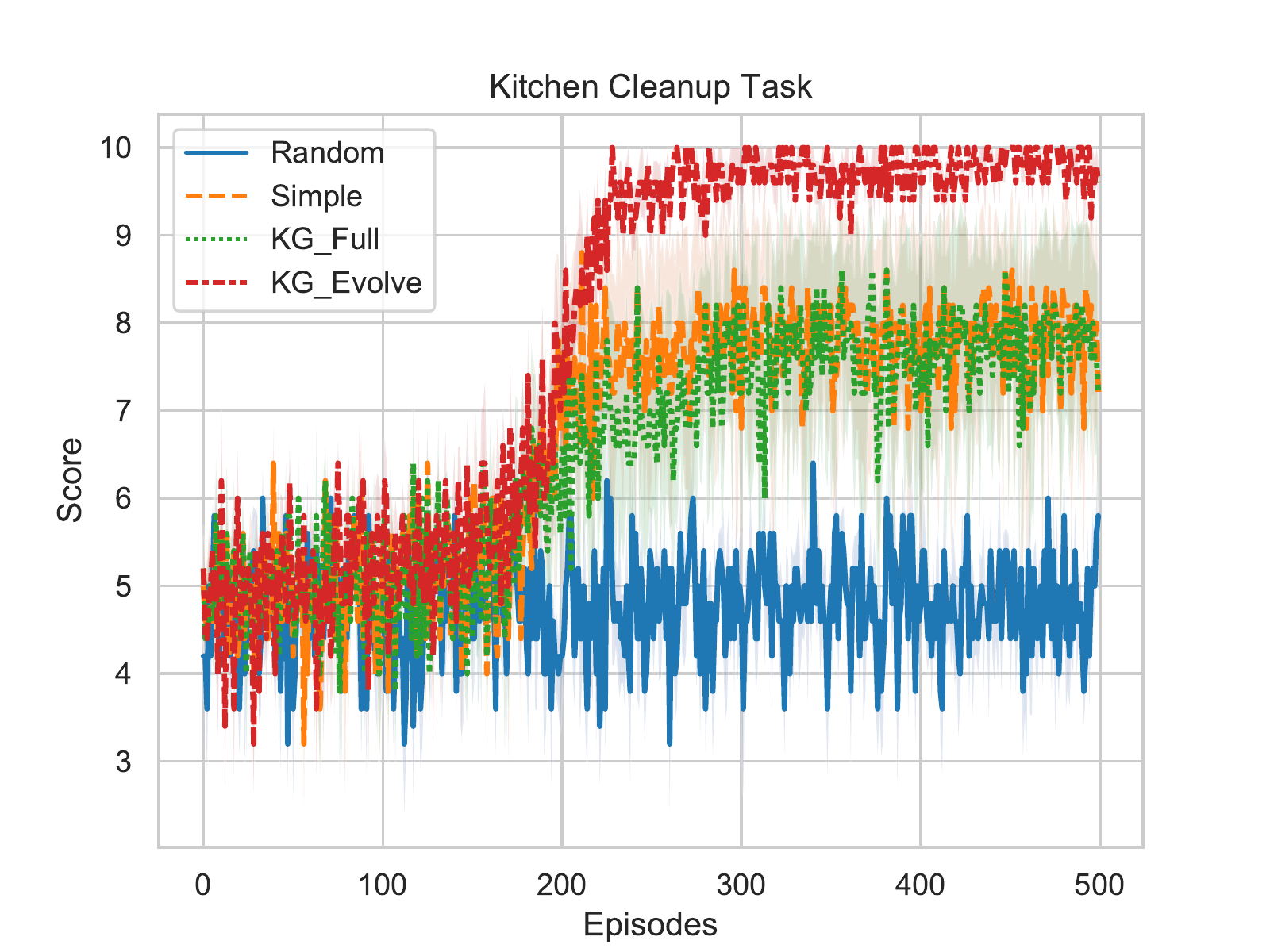}
    \includegraphics[width=0.44\linewidth]{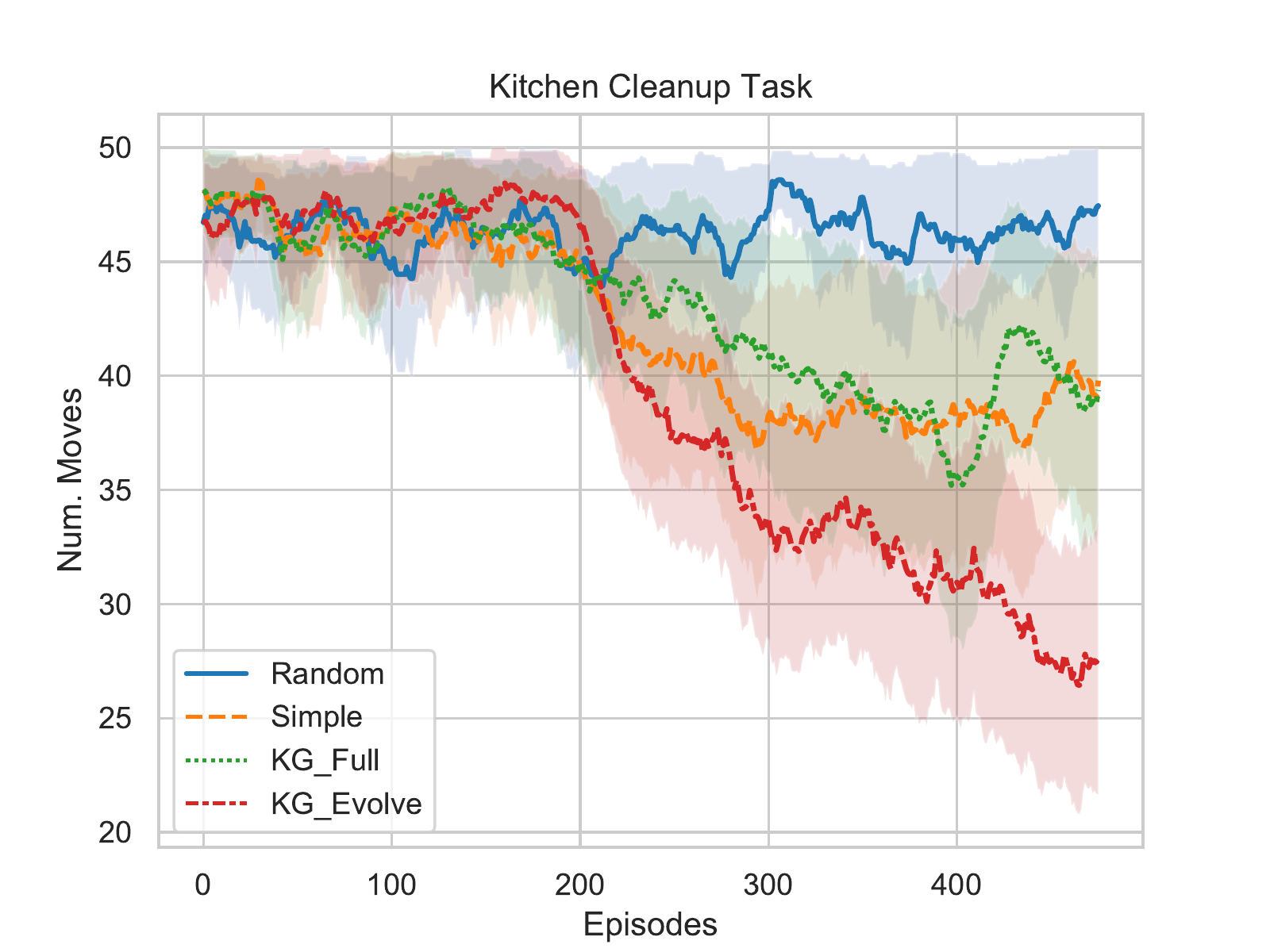}
    \caption{{\bf Simple = ``Text model'', ``KG\_evolve = Text+Commonsense'' in main paper}. Comparison of agents for the Kitchen Cleanup task with and without commonsense knowledge (ConceptNet) with average scores and average moves (averaged over 10 runs).}
    \label{fig:kitchen_scores}
\end{figure*}

\subsection{Results on Kitchen Cleanup}
As explained earlier, we consider two different knowledge-aware agents based on how/when we extract the commonsense subgraph. We consider that the first commonsense agent has access to the entities used in Kitchen cleanup game. Given the entities list, the agent generates a full commonsense subgraph before the start of the game. We call this model \textit{KG\_Full}. Our original model (Text+Commonsense in the main paper) is called \textit{KG\_Evolve} to point that the commonsense subgraph is generated at each time step and evolves over time based on the exploration.

As before, we compare our knowledge-aware RL agents (\textit{KG\_Full} and \textit{KG\_Evolve}) against two baselines for performance comparison: \textit{Random}, where the agent chooses an action randomly at each step; and \textit{Simple} (\textit{Text}-only), where the agent chooses the next action using the text description only and ignores the commonsense knowledge graph. The knowledge-aware RL agents, on the other hand, use the commonsense knowledge graph to choose the next action. The graph is provided in either full-graph setting where all the commonsense relationships between the objects are given at the beginning of the game (\textit{KG\_Full}); or evolve-graph setting where only the commonsense relationship between the objects seen/interacted by the agent until the current steps are revealed (\textit{KG\_Evolve}).  We record the average score achieved by each agent and the average number of interactions (moves) with the environment as our evaluation metrics. Figure~\ref{fig:kitchen_scores} shows the results for the kitchen cleanup task averaged over $5$ runs, with $500$ episodes per run.
\vspace{-1mm}
\subsection{Discussion}
As expected, we see that agents that use the textual description and additionally the commonsense knowledge outperform the baseline random agent. We are also able to demonstrate clearly that the knowledge-aware agent outperforms the simple agent with the help of commonsense knowledge.  The knowledge-aware agent with the evolve-graph setting outperforms both the simple agent as well as the agent with the full-graph setting. We believe that when an agent has access to the full commonsense knowledge graph at the beginning of the game, the agent gets overwhelmed by the amount of knowledge given; and is prone to making noisy explorations in the environment. On the other hand, feeding the commonsense knowledge gradually during the agent's learning process provides more focus to the exploration, and drives it toward the concepts related to the rest of the goals. These results can also be seen as an RL-centric agent-based validation of similar results shown in the broader NLP literature \cite{kapanipathi2020infusing}. We refer the reader to \cite{murugesan2020enhancing} on further discussion on this topic.

\section{Hyperparameters and Complexity}
In addition to the hyperparameters reported in the paper, we used the following settings:



\begin{table}[!h]
\centering
{%
\begin{tabular}{l|l}
\textbf{Hyperparameter}    & \textbf{Setting} \\ \hline
Batchsize                  & 1                \\
Hidden dimension           & 300              \\
Max. \# Steps              & 50               \\
Discount Factor ($\gamma$) & 0.9             
\end{tabular}
}
\caption{Hyperparameters used by the agents}
\end{table}

\noindent The agents were trained in parallel on two machines with the following specifications:

\begin{table}[!h]
\centering
{%
\begin{tabular}{l|l}
\textbf{Resource} & \textbf{Setting}                                   \\ \hline
CPU      & Intel(R) Xeon(R) CPU E5-2690 v4 @ 2.60GHz \\
Memory   & 128GB                                     \\
GPUs     & 2 x NVIDIA Tesla V100 16 GB               \\
Disk1    & 100GB                                     \\
Disk2    & 600GB                                     \\
OS       & Ubuntu 18.04-64 Minimal for VSI.         
\end{tabular}
}
\caption{Resources used by the agents}
\end{table}

\noindent Each agent was trained on a single GPU for approximately 12 hours for the \textit{Text} agent and 16 hours for the \textit{Text + Commonsense} agent for each run.

{
\bibliography{iqa}
}

\end{document}